\documentclass{article}
\usepackage{jmlr2bkk}
\usepackage{amssymb}
\usepackage{amsmath}
\usepackage{amsfonts}
\usepackage{helvet}
\usepackage{color}
\usepackage{courier}

\renewcommand{\eqref}[1]{Eq.\,(\ref{#1})}
\newcommand{\Rew}{\mathcal{V}}
\newcommand{\CC}{\mathcal{C}}
\newcommand{\argmax}{\mbox{\,\rm arg\,max}}
\newcommand{\cc}{\cite}
\newcommand{\tpi}{\tilde{\pi}}
\begin{document}

\title{Asymptotically Optimal Multi-Armed Bandit Policies \\ under a Cost Constraint}
\author{\name Apostolos N. Burnetas \email  aburnetas@math.uoa.gr \\
       \addr   Department of Mathematics\\ National and Kapodistrian  University\\Panepistemiopolis, Athens 15784, Greece\\
        \AND
       \name Odysseas Kanavetas \email okanavetas@sabanciuniv.edu \\
       \addr  Department of Industrial Engineering\\
       Sabanci University\\
       Orhanli
Tuzla, Istanbul 34956, Turkey\\
       \AND
       \name Michael N. Katehakis  \email mnk@rutgers.edu  \\
       \addr Department of Management Science and Information Systems\\
       Rutgers University\\
 100 Rockafeller Rd., Piscataway, NJ 08854,  USA}


\maketitle
\begin{abstract}
\begin{quote}
We develop asymptotically optimal policies for the multi armed bandit (MAB), problem, under a cost constraint. This model is applicable in situations where   each  sample (or activation) from a population (bandit)   incurs a known bandit dependent cost.  Successive samples from each population are iid random variables with unknown distribution.
 The objective is to
  design a feasible policy for deciding from which  population to sample from, so as to maximize the expected sum of outcomes of $n$ total samples or equivalently to minimize the regret due to lack on information on
 sample distributions,
   For this problem we  consider the class of feasible 
   uniformly fast (f-UF) convergent policies,
 that satisfy the cost constraint sample-path wise. We first  establish a necessary  asymptotic  lower bound for   the rate of increase of the regret function of f-UF policies. Then  we construct a class of f-UF    policies and
 provide conditions under which they are   asymptotically optimal within the class of f-UF policies,  achieving this asymptotic lower bound. At the end we provide  the explicit form of such policies for the case in which the
 unknown distributions   are Normal with unknown means and known variances.\end{quote}
\end{abstract}

 
\noindent{\bf \large 1. Introduction}

  Consider the problem of sequential sampling from a finite number of independent statistical populations, where successive samples from a population are iid random variables with unknown distribution.

 Consider the problem of sequential sampling from $k$ independent
statistical populations, $\Pi^i$,  $i=1,\ldots,k$. Successive
samples from population $i$ constitute a sequence of i.i.d. random
variables $X_{1}^{i},X_{2}^{i},\ldots$ following a univariate
distribution with density $f_{i}(\,|\underline{\theta}_i)$ with
respect to a nondegenerate measure $v$. The density $f_i (\,|\,)$ is
known and $\underline{\theta}_i$ is a parameter belonging to some
set $\Theta_i$.
Let
$\underline{\underline{\theta}}=(\underline{\theta}_1,\ldots,\underline{\theta}_k)$
denote the set of parameters,
$\underline{\underline{\theta}}\in\Theta$,  where
$\Theta\equiv\Theta_1\times\ldots\times\Theta_k$.
Given $\underline{\underline{\theta}}$ let
$\underline{\mu}(\underline{\underline{\theta}})=(\mu_1
(\underline{\theta}_1),\ldots,\mu_k (\underline{\theta}_k))$ be the
vector of expected values, i.e. $\mu_i
(\underline{\theta}_i)=E_{\underline{\underline{\theta}}}(X^i)$. The true value  $\underline{\underline{\theta}}_0$  of
$\underline{\underline{\theta}}$ is unknown. We make the
assumption that outcomes from different populations  are independent.

 Sampling from population $\Pi_i$ incurs a positive cost $c^i$  per sample, and  without loss of generality we assume $c^1 \le  c^2 \le \ldots \le  c^N,$
 and not all $c^i$ are equal. The objective is to maximize
the expected average reward per period subject to the constraint that the long-run
average sampling cost per period does not exceed a given upper bound $c^0$ for each
period. Without loss of generality
we assume $c^1\leq c^0< c^k$. In case where $c^0<c^1$, the problem is
infeasible, while in the other case where $c^0\geq c^k$ the cost
constraint is redundant. Let $d=\max\{ j:c^j\leq c^0 \}$. Then
$1\leq d< k$ and $c^d\leq c^0<c^{d+1}$. We consider adaptive
policies which depend only in the past observations of selections
and outcomes.
Specifically, let $A_t,X_t$ , $t=1,2,...$ denote the population
selected and the observed outcome at period $t$. Let $H_t =(A_1
,X_1,....,A_{t-1},X_{t-1})$ denote the history of actions and
observations available at period t.
An adaptive policy is  a sequence $\pi
=(\pi_1,\pi_2,...)$ of history dependent probability distributions on
$\{1,...,k\}$, such that
$ \pi_n (j,h_n)=P(A_n =j|h_n)$ for a given realization $h_n$ of $H_n$.
Given $h_n$, let $T^{\alpha}_{\pi}(n)$ denote the number of
times population $\alpha$ has been sampled during the first n
periods  $T^{\alpha}_{\pi}(n) = \sum_{t=1}^{n}1\{A_t
=\alpha\} $. Let $\Rew_{\pi}(n)$ and $\CC_{\pi}(n)$  be respectively the total
 {\sl  reward earned} and  total {\sl cost incurred}   up to period $n$, i.e.,
\begin{equation}\label{Eq:rew}
 \Rew_{\pi}(n)=
\sum_{i=1}^k     \sum_{t = 1}^{T^i_\pi(n)}  X^i_t,
\end{equation}
\begin{equation}\label{Eq:cost}
\CC_{\pi}(n)=
\sum_{i=1}^k     \sum_{t = 1}^{T^i_\pi(n)}  c^i  .
\end{equation}
 We call an adaptive  policy {\sl feasible} if
\begin{equation}\label{Eq:ccon}
 {\CC_{\pi}(n)}/{n} \leq c^0, \ \ \forall \ n=1,2,\ldots
\end{equation}
 The objective is to obtain  a feasible policy $\pi$     that  maximizes
 in some sense $E_{\underline{\underline{\theta}}} \Rew_{\pi}(n),$
 $ \forall \underline{\underline{\theta}}\in \underline{\underline{\Theta}}.$ In the next section we will show that this is equivalent to minimizing a {\sl regret
 function} $R_{\pi}(\underline{\underline{\theta}},n)$ that represents the expected loss  due to lack of information on the
 sample distributions.     For this,  we  consider the class of feasible policies
 that are    uniformly fast (UF) convergent, in the sense of \cc{bkmab96}; we call these polices (f-UF) policies.
  We first  establish in Theorem 1, a necessary  asymptotic  lower bound for   the rate of increase of the regret function of   f-UF policies. Then  we construct a class of ``block f-UF''    policies and
 provide conditions under which they are   asymptotically optimal within the class of f-UF policies,  achieving this asymptotic lower bound, cf. Theorem 2. At the end we provide  the explicit form of an asymptotically optimal f-UF policy,
   for the case in which the
 unknown distributions   are Normal with unknown means and known variances. These
   policies   form the basis for deriving logarithmic regret polices for more general models, cf. \cc{Auer02b}, \cc{auer2010ucb}, \cc{chk2015},
   \cc{ck2015u}.

The extensive literature on the  multi-armed bandit (MAB) problem, includes the 
following:
\cc{Lai85}, \cc{rmab1995}, \cc{kleinberg2004nearly}, \cc{mahajan2008multi}, \cc{audibert2009exploration}, \cc{auer2010ucb}, \cc{honda2011asymptotically},  \cc{bubeck2012best}, \cc{cowan15s} and references therein.  As far as we know, the    first formulation of the MAB problem with a side constraint considered herein was given  in \cc{bkmab98}.
\cc{tran2010epsilon}, considered the problem when the cost of activation of each arm is fixed and becomes known after the arm is used once.
 \cc{bu+ka12} considered a version
of this problem   and constructed a consistent  policy (i.e.,
 with regret $R_\pi(n)=o(n)$).  In the present paper we employ  a stricter version of the average cost constraint that
requires the average sampling cost not to exceed $c^0$ at any time
period and not only in the limit. \cc{badanidiyuru2013bandits}, considered the problem where there can be more than one side constraints (``knapsack'')
 and showed how to construct polices with sub-linear  regret. They also discuss
 interesting applications of the model, such as  to: problems of dynamic pricing
 \cc{wang2014close}, \cc{johnson2015online}, dynamic procurement \cc{singla2013truthful}, and auctions \cc{tran2014efficient}.
 \cc{ding2013multi} constructed
 UF policies (i.e.,  with regret $R_\pi(n)=o(\log n)$) for cases in which
activation costs are bandit dependent iid random variables. For other recent
related work we refer to: \cc{guha2007approximation}, \cc{tran2012}, \cc{thomaidou2012toward}, \cc{lattimore2014optimal}, \cc{sen2015adaptive}. 

 For  other    work in this area   we refer to   \cite{Kat86}, \cite{Kat87}, \cite{burnetas1993sequencing}, \cite{BKlarge1996}, \cite{lagoudakis2003least},
\cite{bartlett2009regal}, \cite{tekin2012approximately}, \cite{jouini2009multi},
 \cite{dayanik2013asymptotically}, \cite{filippi2010optimism}, \cite{osband2014near}. As well as \cc{burnetas2003asymptotic},  \cite{audibert2009exploration},   
\cite{auer2010ucb}, \cite{gittins2011multi}, \cite{bubeck2012best},  
 \cite{cappe2013kullback}, \cite{kaufmann14}, 
\cite{2014minimax},  
 \cite{cowan15s}, \cite{cowan2015multi}, 
and references therein. 
    For     dynamic programming extensions 
we refer to  
\cite{bkmdp97}, \cite{butenko2003cooperative}, \cite{optimistic-mdp}, \cite{audibert2009exploration}, \cite{littman2012inducing}, \cite{feinberg2014convergence} and references therein. 
\\

\noindent{\bf \large 2. Model description - Preliminaries}

The complete information problem where $\underline{\underline{\theta}}$ is known, the expected average reward is to be maximized, and the expected average cost does not exceed $c^0$, can be solved via
 the following  linear program (LP-1) which is instrumental in the development of the lower bounds and the asymptotically optimal policy.
\begin{eqnarray}
\text{(LP-1):} \  \ z^{*}(\underline{\underline{\theta}}) & = & \max
\sum_{j=1}^{k}\mu_{j}(\underline{\theta}_j) x_j \nonumber\\
&&\sum_{j=1}^{k}c^j x_j  +y =   c^0 \label{lp}\\
&&\sum_{j=1}^{k}x_j =1 \nonumber\\
&&x_j \geq 0,\forall j\ y \geq 0. \nonumber
\end{eqnarray}
The solution is a randomized sampling policy which at each period
selects population $j$ with probability $x_j$, for $j=1,\ldots,k$,
where the randomization probabilities $x_j$ are an optimal solution
to the above linear program (LP), cf. \cc{bu+ka12,bkmab98}.
However, such  policy may not be feasible in our framework that requires
${\CC_{\pi}(n)}/{n} \leq c^0, \ \forall \ n=1,2,\ldots,$
because simple randomization
may lead to  sampling in such a way that ${\CC_{\pi}(n)}/{n}$  exceeds
$c^0$, for some    periods. However, in the complete
information setting,  under the assumption  that the
 coefficients $c^j$ are all rational, any optimal solution of LP-1
which is an extreme point is also rational, thus an optimal  randomized
policy can be implemented as a {\sl periodic sampling policy} within {\sl blocks}
of time periods within which  the order of sampling can be set so that the
sampling cost constraint is never violated, and the sampling frequencies
remain equal to   $x_j$.  We use generalizations of this idea in the
incomplete information framework in the sequel.

We next introduce necessary notation regarding the LP-1.
First, its  dual problem (DLP-1)   is

\begin{eqnarray*}
\text{(DLP-1):} \  \ z^{*}_{D}(\underline{\underline{\theta}}) & = & \min \ g+c^0 \lambda
\\ && g+c^1 \lambda \geq \mu_1 (\underline{\theta}_1) \\ &&\hspace{1cm}
\vdots \label{dlp}\\
&& g+c^k \lambda \geq \mu_k (\underline{\theta}_k) \\\ && g\in
\textbf{R},\lambda \geq 0 .
\end{eqnarray*}

A basic matrix $B$ is of the form $\left( \begin{array}{cc} c^i &
c^j\\ 1 &
1\\
\end{array} \right)$, for some $i\leq d < j$ or $\left( \begin{array}{cc} c^i & 1\\ 1 & 0
\\
\end{array} \right)$ for some $i\leq d$. They  correspond to sampling from the pair $(i,j)$
or population $i$, respectively. We denote the Basic Feasible Solution (BFS)
corresponding to matrix $B$ as $b=\{i,j\}$ or $b=\{i\}$, respectively.
Note that in the case of degenerate BFS $b$, more than one matrices $B$ correspond to the same $b$.

 We use $F$ to denote the set of BFS:
\begin{equation}
  F=\{ b \ : \ b=\{i,j\}, \ i\leq d\leq j \mbox{ or }b=\{i\}, \ i\leq d  \}.\nonumber
\end{equation}
\noindent Since the feasible region of \eqref{lp} is bounded, $F$ is a finite set.

For a basic matrix $B$, let $v^B= (\lambda^B, g^B)$ denote the dual
vector corresponding to $B$, i.e., $v^B = \mu_B
(\underline{\underline{\theta}}) B^{-1}$, where $\mu_B
(\underline{\underline{\theta}})=(\mu_i (\underline{\theta}_i),
\mu_j (\underline{\theta}_j))$, or $\mu_B
(\underline{\underline{\theta}})=(\mu_i (\underline{\theta}_i), 0)$,
depending on the form of $B$.

Regarding optimality, a BFS is optimal if and only if for at least one
corresponding basic matrix $B$ the reduced costs (dual slacks) are all
nonnegative:
\begin{equation}
  \label{eq:phidef}
\phi^B_{\alpha} (\underline{\underline{\theta}}) \equiv c^{\alpha}
\lambda^B + g^B - \mu_{\alpha}(\underline{\theta}_{\alpha})
  \geq 0, \ \alpha=1,\ldots, k.\nonumber
\end{equation}

A basic matrix $B$ satisfying this condition is optimal.
%
It is easy to show that the reduced cost can be expressed as a
linear combination of the unknown population means, i.e.,
$\phi_{\alpha}^{B}(\underline{\underline{\theta}})=\underline{w}_{\alpha}^{B}\underline{\mu}(\underline{\underline{\theta}})$,
where $\underline{w}_{\alpha}^{B}$ is an appropriately defined
vector that does not depend on
$\underline{\mu}(\underline{\underline{\theta}})$.
In the sequel we use the notation $s(\underline{\underline{\theta}})$ to denote
  the set with optimal solutions of LP-1 for a
vector $\underline{\mu}(\underline{\underline{\theta}})$, i.e.,
$
s(\underline{\underline{\theta}})=\{ b \in F : b \mbox{ corresponds
to an optimal BFS} \}.
$

We define the loss or regret
function of policy $\pi$ as the finite horizon loss in expected
reward with respect to the optimal policy under complete
information:
\begin{eqnarray}
R_{\pi}(\underline{\underline{\theta}},n)&=&
nz^{*}(\underline{\underline{\theta}})-
E_{\underline{\underline{\theta}}} \Rew_{\pi}(n) \nonumber\\
&=&n z^*(\underline{\underline{\theta}})-
\sum_{j=1}^{k}\mu_j (\underline{\theta}_j) E_{\underline{\underline{\theta}}}T_{\pi}^{j}(n)\label{reg1}
\end{eqnarray}

We next derive an equivalent expression that relates the regret to
the solution of the complete information LP. Recall that for any
basic matrix $B$ which corresponds to an optimal solution of LP-1,
from the   DLP-1 program  we have that $\forall j$:
$
z^*(\underline{\underline{\theta}})=c^0 \lambda^B +g^B \mbox{ and
}\mu_j(\underline{\theta}_j)=c^j \lambda^B
+g^B-\phi_{j}^{B}(\underline{\underline{\theta}}).$ These relations and \eqref{reg1} imply:
\begin{equation}\label{reg2}
R_{\pi}(\underline{\underline{\theta}},n)=\sum_{j=1}^{k}\phi_{j}^{B}(\underline{\underline{\theta}})
E_{\underline{\underline{\theta}}}T_{\pi}^{j}(n)+\lambda^B \
\sum_{j=1}^{k}(c^0 -
c^j)E_{\underline{\underline{\theta}}}T_{\pi}^{j}(n),
\end{equation}
for any $\underline{\underline{\theta}}\in\Theta$.
%
%

  We now state:


{\bf Definition 1.}  a)
 A  feasible policy $\pi$ is called {\sl consistent}       if
\begin{equation}
R_{\pi}(\underline{\underline{\theta}},n)=o(n), \
n\rightarrow\infty, \    \forall \
\underline{\underline{\theta}}\in\Theta.\nonumber
\end{equation}

b)     A  feasible policy $\pi$ is called {\sl f-uniformly fast} (f-UF)  if
\begin{equation}
R_{\pi}(\underline{\underline{\theta}},n)=o(n^{a}), \
n\rightarrow\infty, \ \forall \ a>0, \ \forall \
\underline{\underline{\theta}}\in\Theta.\nonumber
\end{equation}
In the sequel we will show that    there exist f-UF policies, following the
approach of \cite{bkmab96},  by construction   of a function $M(\underline{\underline{\theta}})$ and a f-UF policy $\pi^0 $ such
that $$\liminf R_{\pi^0}(\underline{\underline{\theta}},n)/\log n \leq
M(\underline{\underline{\theta}})\ \ \forall
\underline{\underline{\theta}}\in\Theta.$$ As we will be shown later, (Theorem 1), policy $\pi^0$ has the much stronger property of asymptotic optimality. Indeed, $M(\underline{\underline{\theta}})$ is also a uniform lower bound on the limit of $R_{\pi}(\underline{\underline{\theta}},n)/\log n$, of any f-UF policy.\\

\noindent{\bf \large 3. Lower Bound for the Regret}

 Recall that for $b\in F$, $b$ is
an optimal solution of linear program LP-1 for some
$\underline{\underline{\theta}}\in\Theta$ if and only if for at
least one corresponding basic matrix $B$, $\phi^B_{\alpha}
(\underline{\underline{\theta}})\geq 0, \ \alpha=1,\ldots, k$.

For any $b\in s(\underline{\underline{\theta}})$, where $b=\{i,j\}$
or $\{i\}$ and $\alpha\neq i,j,$
 we  define the sets  $\Delta\Theta_{\alpha}(\underline{\underline{\theta}})$  and $D(\underline{\underline{\theta}})$,
as follows. The first set includes all perturbed values $\underline{\theta}_{\alpha}^{'}$
of $\underline{\theta}_{\alpha}$ of population $\alpha$, such that the complete info problem under $\underline{\underline{\theta}}^{'}$ where only $\underline{\theta}_{\alpha}$ is perturbed to $\underline{\theta}_{\alpha}^{'}$ has a unique optimal BFS which includes population $\alpha$.
The second set  $D(\underline{\underline{\theta}})$, contains all
  populations which are not contained in any optimal solution
under parameter set $\underline{\underline{\theta}}$ but, by varying
only parameter $\underline{\theta}_{\,\alpha}$, a uniquely optimal BFS
that contains them can be found. Formally,
\begin{equation}
\Delta\Theta_{\alpha}(\underline{\underline{\theta}})= \{
\theta_{\alpha}^{'}\in\Theta_{\alpha}:
s(\underline{\underline{\theta}}^{'})=\{\{i,\alpha\}\mbox{ or
}\{\alpha,j\}\mbox{ or }\{\alpha\}\} \},\nonumber
\end{equation}
where $\underline{\underline{\theta}}^{'}=(\underline{\theta}_1,
\ldots,\underline{\theta}_{\alpha}^{'},
\ldots,\underline{\theta}_k)$, is a new vector such that only
parameter $\underline{\theta}_{\alpha}^{'}$ is changed from
$\underline{\theta}_{\alpha}$. Then, $D(\underline{\underline{\theta}})$ is the set of populations, which are not optimal under $\underline{\underline{\theta}}$, but become part of a uniquely optimal BFS after a parameter change of $\underline{\theta}_{\alpha}$ only. 
\begin{equation}
D(\underline{\underline{\theta}})=\{ \alpha: \ \alpha \notin b
\mbox{ for any } b\in s(\underline{\underline{\theta}}) \mbox{ and }
\Delta\Theta_{\alpha}(\underline{\underline{\theta}})\neq\emptyset
\}, \nonumber
\end{equation}
Let
$I(\underline{\theta}_{\alpha},\underline{\theta}_{\alpha}^{'})$ denote
the Kullback-Leibler information number, defined as
\begin{equation}
I(\underline{\theta}_{\alpha},\underline{\theta}_{\alpha}^{'})=\int_{-\infty}^{+\infty}
\log
\frac{f(x;\underline{\theta}_{\alpha})}{f(x;\underline{\theta}_{\alpha}^{'})}
f(x;\underline{\theta}_{\alpha})dv(x).\nonumber
\end{equation}

 Now we can define the minimum deviation, in the sense of the
Kullback-Leibler information number, of parameter
$\underline{\theta}_{\alpha}^{'}$ from $\underline{\theta}_{\alpha}$
in order to achieve that population $\alpha$ becomes optimal
under $\underline{\theta}_{\alpha}^{'}$.
\begin{eqnarray}
K_{\alpha}(\underline{\underline{\theta}})&=&
\inf\{ I(\underline{\theta}_{\alpha},\underline{\theta}_{\alpha}^{'}): \ \underline{\theta}_{\alpha}^{'}\in\Delta\Theta_{\alpha}
(\underline{\underline{\theta}}) \} .\nonumber
\end{eqnarray}
We have:

\noindent
{\bf Lemma 1}
For any $\underline{\underline{\theta}}$, and any optimal matrix $B$ under $\underline{\underline{\theta}}$,
$\exists$ $\rho=\rho(\underline{\underline{\theta}},\alpha,B)>0$ such that for any
$\underline{\theta}_{\alpha}^{'}\in\Delta\Theta_{\alpha}(\underline{\underline{\theta}}):$\\
$\textbf{(i)}$ $\phi^B_{j}
(\underline{\underline{\theta}}^{'})=\phi^B_{j}
(\underline{\underline{\theta}})\geq 0, \ \forall \ j\neq\alpha $
and
$\phi^B_{\alpha} (\underline{\underline{\theta}}^{'})=\phi^B_{\alpha} (\underline{\underline{\theta}})
+\mu_{\alpha}(\underline{\theta}_{\alpha})-\mu_{\alpha}(\underline{\theta}_{\alpha}^{'})<0$,\\
$\textbf{(ii)}$
$\mu_{\alpha}^*(\underline{\underline{\theta}})<\mu_{\alpha}(\underline{\theta}_{\alpha}^{'})<\mu_{\alpha}^*(\underline{\underline{\theta}})+\rho$,
where 
$\mu_{\alpha}^*(\underline{\underline{\theta}})=\phi^B_{\alpha}
(\underline{\underline{\theta}})
+\mu_{\alpha}(\underline{\theta}_{\alpha})$.

 The above Lemma implies the following
form for $K_{\alpha}(\underline{\underline{\theta}})$ which  is necessary
for the proof of Lemmas and Theorems of the paper:  
$$
K_{\alpha}(\underline{\underline{\theta}})= \inf\{
I(\underline{\theta}_{\alpha},\underline{\theta}_{\alpha}^{'}): \
\underline{\theta}_{\alpha}^{'}\in\Theta_{\alpha}, \
\mu_{\alpha}^*(\underline{\underline{\theta}})<\mu_{\alpha}(\underline{\theta}_{\alpha}^{'})<\mu_{\alpha}^*(\underline{\underline{\theta}})+\rho\},
$$
where $\rho=\rho(\underline{\underline{\theta}},\alpha,B)>0.$

  Lemma $2$ and Proposition $1$ below  are used  to establish
   the following Lemma $3$ from which Theorem 1  for the
regret function follows.

First note that in \eqref{reg2} both terms are nonnegative, the
first because of optimality and the second because of feasibility.
Therefore, a necessary and sufficient condition for a
policy $\pi$ to be f-UF  is that for
$\underline{\underline{\theta}}\in\Theta$ and any optimal BFS $b$
under $\underline{\underline{\theta}}$ and for all B corresponding to b:

\begin{equation}
\phi_{j}^{B}(\underline{\underline{\theta}})\lim_{n\rightarrow\infty}\frac{E_{\underline{\underline{\theta}}}T_{\pi}^{j}(n)}{n^a}=0,
\mbox{ for all }  a>0, \ j\notin b,\label{ug1}
\end{equation}
and
\begin{equation}
\lambda^{B}\lim_{n\rightarrow\infty}\frac{\sum_{j\in b}(c^0 -
c^j)E_{\underline{\underline{\theta}}}T_{\pi}^{j}(n)}{n^a}=0.\label{ug2}
\end{equation}

We can now state:

\noindent{\bf Lemma 2} Assume a uniquely optimal BFS and $B\in
s(\underline{\underline{\theta}})$. Then\\ \emph{\textbf{(i)}} if
$B=\left(
\begin{array}{cc} c^i & c^j\\ 1 &
1\\
\end{array} \right)$, for some $i\leq d < j$ then $\lambda^B >0$,\\
\emph{\textbf{(ii)}} if $B=\left( \begin{array}{cc} c^i & 1\\ 1 & 0
\\
\end{array} \right)$, for some $i\leq d$ then $\lambda^B =0$.

\noindent{\bf Proposition 1} For any f-UF policy $\pi$ and for all
$\underline{\underline{\theta}}\in\Theta$ we have that for
$\alpha\in D(\underline{\underline{\theta}})$, any
$\underline{\underline{\theta}}^{'}\in\Delta(\underline{\underline{\theta}})$
and for all positive sequences:  $\beta_n=o(n)$ it is true that
\begin{equation}
P_{\underline{\underline{\theta}}^{'}}[T^{\alpha}_{\pi}(n)
<\beta_n]=o(n^{a-1}), \mbox{ for all }a>0.\nonumber
\end{equation}

So far we have shown that a necessary condition for a uniformly fast
policy is that $\forall \ \underline{\underline{\theta}}\in
\underline{\Theta}$, and $\forall \ \alpha\in
D(\underline{\underline{\theta}})$ it must be true that the number
of samples from populations $j_0$ and $\alpha$ are at least
$\beta_n$ correspondingly, because
$P_{\underline{\underline{\theta}}^{'}}(T^{j_0}_{\pi}(n)\leq
\beta_n)=o(n^{a-1})$,
$P_{\underline{\underline{\theta}}^{'}}(T^{\alpha}_{\pi}(n)\leq
\beta_n)=o(n^{a-1})$ for any positive sequence of constants $\beta_n
=o(n)$.

\noindent{\bf  Lemma 3} If
$P_{\underline{\underline{\theta}}^{'}}[T^{\alpha}_{\pi}(n)
<\beta_n]=o(n^{a-1}),$ for all $a>0$ and a positive sequence
$\beta_n=o(n)$ then
\begin{equation}
\lim_{n\rightarrow\infty}P_{\underline{\underline{\theta}}}[T^{\alpha}_{\pi}(n)<\frac{\log
n}{K_{\alpha}(\underline{\underline{\theta}})}]=0,\nonumber
\end{equation}
for all $\underline{\underline{\theta}}\in\Theta$ and
$\alpha\in\Delta(\underline{\underline{\theta}})$.\\

We next define the function $M(\underline{\underline{\theta}}) $ and prove the main theorem of this section. Let
$$
M(\underline{\underline{\theta}}) =\sum_{j\in
D(\underline{\underline{\theta}})}\frac{\phi_{j}^{B}(\underline{\underline{\theta}})}{K_j(\underline{\underline{\theta}})}.
$$

\noindent{\bf Theorem 1}
If $\pi$ is an f-UF policy then
\begin{equation}
\liminf_{n\rightarrow\infty}\frac{R_{\pi}(\underline{\underline{\theta}},n)}{\log
n}\geq M(\underline{\underline{\theta}}),
\ \forall  \underline{\underline{\theta}}\in\Theta.\nonumber
\end{equation}
\noindent
{\bf Proof} Recall,
\begin{equation}
R_{\pi}(\underline{\underline{\theta}},n)=\sum_{j=1}^{k}\phi_{j}^{B}(\underline{\underline{\theta}})
E_{\underline{\underline{\theta}}}T^{j}_{\pi}(n)+\lambda^B [ nc^0-
E_{\underline{\underline{\theta}}}C_{\pi}(n) ], \nonumber
\end{equation}
and by Lemma 3, using the Markov inequality, we obtain that if $\pi$
is f-UF, then
\begin{equation}
\liminf_{n\rightarrow
\infty}\frac{E_{\underline{\underline{\theta}}}T^{j}_{\pi}(n)}{\log
n}\geq\frac{1}{K_{j}(\underline{\underline{\theta}})}, \  \forall   j\in D(\underline{\underline{\theta}}), \ \forall  \underline{\underline{\theta}}\in\Theta.\nonumber
\end{equation}

Also, we have from Lemma 2 that $\lambda^B\geq0$ and from Eq.
(\ref{Eq:ccon}), we have  that $nc^0-
E_{\underline{\underline{\theta}}}\CC_{\pi}(n)\geq 0$, for all $n$.
Finally, we have that the optimal populations under
$\underline{\underline{\theta}}$ have
$\phi_{j}^{B}(\underline{\underline{\theta}})=0$, thus
\begin{equation}
\liminf_{n\rightarrow\infty}\frac{R_{\pi}(\underline{\underline{\theta}},n)}{\log
n}\geq\sum_{j\in
D(\underline{\underline{\theta}})}\frac{\phi_{j}^{B}(\underline{\underline{\theta}})}{K_j(\underline{\underline{\theta}})},
\mbox{ for all } \underline{\underline{\theta}}\in\Theta.\nonumber
\end{equation}

\noindent{\bf \large 4. Blocks and Block Based Policies}

We consider a class of policies such that
sampling is performed in groups of subsequent periods called
sampling blocks, of finite length, where the total cost of actions
in each block satisfies the cost constraint of \eqref{Eq:ccon}
as follows. Define the differences

\begin{equation}\label{Eq:delta}
\delta^i \equiv c^i -c^0.\nonumber
\end{equation}
$\delta^i$ expresses the net cost effect of a single observation from a
population $i$ on the sampling budget. This effect is a net cost if
$\delta^i>0$ or net savings if $\delta^i<0$.

The original problem is equivalent to the transformed problem where
$c^{i}=\delta^i$, $i=1,...,k$, $c^{0} =0$ and the sampling
constraint is

\begin{equation} \frac{1}{n}\sum_{t=1}^{n}\delta^{A_t} \leq 0, \
\forall \ n.\nonumber
\end{equation}

Since $\delta^i $ are assumed to be rational, for each
$i=1,\ldots,k$ and there is a finite number of them we may assume,
without loss of generality, that they are all integers.

 Let $J\subseteq \{ 1,...,k \}$ be the subset of populations sampled
within a sampling block.
The ``cheap'' populations in $J$ must be sampled often enough to
finance sampling of the ``expensive'' ones. Mathematically it
suffices to find $\{ m_j , j\in J \}$ such that each population $j
\in J$ is sampled $m_j$ times, and $\sum_{j \in J} m_j \delta^j \leq
0$, $m_j \in \textbf{N}$, $\forall \ j \in J$. Any block with $m_j$
satisfying the previous properties is called admissible. One
possibility is to consider the smallest block, which will be
appropriate in the incomplete information case. Thus the
minimum length of the sampling block, $\ell(J),$ is the solution of
the following integer linear program
$$
\ell(J) =  \min \{ \sum_{j\in J} m_j  \ : \ \sum_{j\in J}m_j
\delta^j \leq 0\ \& \ m_j \in \textbf{N}, \ \forall \ j\in J \}.
$$
%

An optimal solution of LP-1 specifies randomization probabilities
that guarantee maximization of the average reward subject to the
cost constraint. The populations into this optimal solution define
the set $J$, and $J$, $\delta^i$ and $\ell$ are observable
constants.

We use the  Initial Sampling Block (ISB) and Linear Programming Block (LPB) blocks below  to define a class of policies $\tilde{\pi}$
that  are feasible, as follows.

a) A policy $\tilde{\pi}$ starts with an ISB   block during which
all populations $\{ 1,...,k \}$ are
sampled at least a predetermined number of times $n_0$,
 with  a sufficient    number of samples taken
 from cheap (small $c^i$) populations,
so that the   constraint of \eqref{Eq:ccon} is satisfied
sample path-wise.  This block is necessary in order to  obtain initial
 estimates of
$\mu_j (\underline{\theta}_j)$ for all populations. This means
that the ISB block has the minimum length of $\ell(J)$,  defined
above, with $J=\{ 1,...,k \}$.

b) After a completion of an ISB block a   $\tilde{\pi}$  policy chooses
any   BFS (or equivalently a single population $\{ i \}$ or a pair of
$\{ i,j \}$)  and continues sampling for a block of time periods LPB=LPB(b)
as  follows.

i) When $b=\{ i \}$, (which means that   $c^i\leq c^0$) $\tilde{\pi}$
samples from population $i$ only once. In this case we define the
LPB  block to have  length equal to:  $m_{i}^{b}=1$, and its sampling
  frequency $x_{i}$ to be equal to $1$, $x_{i}=1$.

ii) When  $b=\{ i,j \}$, $\tilde{\pi}$  samples a number of times   each population in $\{ i,j \}$ in $b$ so as the cost feasibility of $\tilde{\pi}$
is maintained during the block. The latter is accomplished by taking the
length of the LPB  block to be equal to: $m_{i}^{b}+m_{j}^{b}=|\delta^{j}|+|\delta^{i}|$, where $m_{i}^{b}=|\delta^{j}|$
and $m_{j}^{b}=|\delta^{i}|,$
and sampling the least cost population first in such a way that the frequencies are equal to the randomization probabilities:
\begin{equation}\label{Eq:xij}
x_{i}= \frac{|\delta^{j}|}{|\delta^{i}|+|\delta^{j}|},
 \ x_{j}=
 \frac{|\delta^{i}|}{|\delta^{i}|+|\delta^{j}|},
  \nonumber
\end{equation}

\noindent{\bf Remark 1} Note that in  the second  case of an LPB,
the randomization probabilities for $\{i,j\}$,  and the   block
length $m_{i}^{b}+m_{j}^{b}$,
 are computed
without solving LP-1, using the known, cf. \eqref{Eq:delta},
$\delta$'s.

  Note that a {\sl block based policy} is a
well defined adaptive policy.
In the sequel we restrict our attention to  {\sl block based policies}; for notational
simplicity we will simply write $\pi$ in place of $\tpi$, when there is
no risk for confusion.

Assume that we have $l$ successive blocks  we take
$\widetilde{T}_{\pi}^b(l)$ to be  the number of LPB($b$) type blocks
  in first $l\geq 2$ blocks (since   for $l=1 $ we start with an ISB block).
Thus $\sum_{b\in K}\widetilde{T}_{\pi}^b(l)=l-1$. Let $S_{\pi}(l)$ be
the total length of first $l$ blocks and let $L_n=L_{\tpi}(n)$ denote the number of blocks
in n periods. We can easily show that
\begin{equation}
T_\pi^\alpha(S_\pi(l))=\sum_{b:\alpha\in
b}m_{\alpha}^{b}\,\widetilde{T}_\pi^b(l)+m_{\alpha},\nonumber
\end{equation}
where $m_{\alpha}^{b}$ is the number of samples from population
$\alpha$ between a LPB$(b)$ and $m_{\alpha}$ is the number of
samples from population $\alpha$ in the ISB block. Now we can define the regret of blocks
\begin{eqnarray}
\widetilde{R}_{\pi}(\underline{\underline{\theta}},l)= z^{*}(\underline{\underline{\theta}}) \
E_{\underline{\underline{\theta}}}S_\pi(l)\,
-
E_{\underline{\underline{\theta}}}\sum_{j=1}^{k}\sum_{b\in
K}\mu_{j}(\underline{\theta}_j)\,
m_{j}^{b}\,\widetilde{T}_\pi^b(l)
\nonumber\\
 \
-\sum_{j=1}^{k}
\mu_{j}(\underline{\theta}_j)m_{j}.\nonumber
\end{eqnarray}

We note that
\begin{equation}
T_\pi^\alpha(S_\pi(L_n)) \leq T^{\alpha}_{\pi}(n)  \leq
T_\pi^\alpha(S_\pi(L_n))+M_{\alpha},\label{as3}
\end{equation}
where $M_{\alpha}$ is the maximum number of times where population
$\alpha$ appears in every block. Thus we obtain  the following
relation for the two types of regret,
\begin{eqnarray}
\widetilde{R}_{\pi}(\underline{\underline{\theta}},L_n)
+(n-E_{\underline{\underline{\theta}}}S_\pi(L_n))\, z^*
(\underline{\underline{\theta}})
- \sum_{j=1}^{k}M_j \, \mu_{j}(\underline{\theta}_j)\nonumber\\
\leq R_{\pi}(\underline{\underline{\theta}},n) \leq
\widetilde{R}_{\pi}(\underline{\underline{\theta}},L_n) +
(n-E_{\underline{\underline{\theta}}}S_\pi(L_n))\, z^*
(\underline{\underline{\theta}}).\label{as1}
\end{eqnarray}
The above and \eqref{as1} imply the following relation between the
two regret functions,

\begin{equation}
\limsup_{n\rightarrow\infty}\frac{R_{\pi}(\underline{\underline{\theta}},n)}{\log
n}
=\limsup_{n\rightarrow\infty}\frac{\widetilde{R}_{\pi}(\underline{\underline{\theta}},L_n)}{\log
L_n}.\label{as2}
\end{equation}
From \eqref{as2}, it follows that
 in order to find a policy that achieves the
lower bound for $R_{\pi}(\underline{\underline{\theta}},n)$, it suffices to   find a policy that
 achieves the
lower bound for $\widetilde{R}_{\pi}(\underline{\underline{\theta}},L_n)$.\\

\noindent{\bf \large 5. Asymptotically Optimal Policies}

In this section we provide a general method to construct
asymptotically optimal policies $\pi^0$ that achieve the lower bound
for the regret. To state the policy we need some definitions. We
define at any block $l$ and for every population $\alpha$ as
$\widetilde{\mu}_{\alpha}$
\begin{equation}
\widetilde{\mu}_{\alpha}=\sup_{\underline{\theta}_{\alpha}^{'}}\{
\mu_{\alpha}(\underline{\theta}_{\alpha}^{'}):I(\hat{\underline{\theta}}_{\alpha}^{l},
\underline{\theta}_{\alpha}^{'})\leq\frac{\log
S_\pi(l-1)}{T_\pi^\alpha(S_\pi(l-1))} \},\nonumber
\end{equation}
and as
$\Phi_{l}^{(\hat{B},\hat{\underline{\underline{\theta}}}^{l})}$
\begin{equation}
\Phi_{l}^{(\hat{B},\hat{\underline{\underline{\theta}}}^{l})}=\{
\alpha \ : \mu_{\alpha}^*(\hat{\underline{\underline{\theta}}}^{l})
< \widetilde{\mu}_{\alpha} <
\mu_{\alpha}^*(\hat{\underline{\underline{\theta}}}^{l}) +
\rho(\hat{\underline{\underline{\theta}}}^{l},\alpha,\hat{B})
\}.\nonumber
\end{equation}
We recall that if we have an optimal BFS $b$, where $b=\{i,j \} $ or
$\{i\}$ then the optimal solution is $z^b=\mu_i x_i+\mu_j x_j$ or
$z^b=\mu_i$.\\

\noindent
\textbf{INFLATED Z-POLICY $\pi^0$:}\\

  Start with   one ISB block in
order to have at least one estimate from each population. Then,

\noindent
{\bf Step 1} Assume that at the beginning of block $l$, $l>1$, we
have the estimates $\hat{\underline{\underline{\theta}}}^{l}$, from
the previous $l-1$ blocks  with
$\mu_1 (\hat{\underline{\theta}}_{1}^{l}),...,\mu_k
(\hat{\underline{\theta}}_{k}^{l})$.
We take the solution of LP-1:
\begin{equation}
z^{b(\hat{\underline{\underline{\theta}}}^{l})}=
\max_{\widetilde{b}_i (\hat{\underline{\underline{\theta}}}^{l})}\{
z^{\widetilde{b}_i (\hat{\underline{\underline{\theta}}}^{l})}: \
\widetilde{T}_\pi^{\widetilde{b}_i
(\hat{\underline{\underline{\theta}}}^{l})}(l)\geq \tau(l-1)
\}\nonumber
\end{equation}
where $\widetilde{b}_{i}$ are all the BFS in $F$ and $\tau $ is any
fixed constant  in:  $(0,1/|F|)$.

\noindent
{\bf Step 2} Then for every $\alpha=\{ 1,\ldots,k \}$, we compute
the $\widetilde{\mu}_{\alpha}$'s and
$\Phi_{l}^{(\hat{B},\hat{\underline{\underline{\theta}}}^{l})}$'s.

Then, if
$\Phi_{l}^{(\hat{B},\hat{\underline{\underline{\theta}}}^{l})}=\emptyset$,
we take   $\pi^{0}(\hat{\underline{\underline{\theta}}}^{l})=b(\hat{\underline{\underline{\theta}}}^{l})$),
otherwise for every $\alpha \in
\Phi_{l}^{(\hat{B},\hat{\underline{\underline{\theta}}}^{l})}$ we
define the index:
\begin{equation}
u_{\alpha}(\hat{\underline{\underline{\theta}}}^{l},\underline{\theta}_{\alpha}^{'})=
\max_{\underline{\theta}_{\alpha}^{'}}\{
z^{b_{\alpha}(\hat{\underline{\underline{\theta}}}^{l},\underline{\theta}_{\alpha}^{'})}:
I(\hat{\underline{\theta}}_{\alpha}^{l},\underline{\theta}_{\alpha}^{'})\leq\frac{\log
S_\pi(l-1)}{T_\pi^{\alpha}(S_\pi(l-1))} \},\nonumber
\end{equation}
and we take
\begin{equation}
\pi^{0}(\hat{\underline{\underline{\theta}}}^{l})=\argmax_{\ }
\{
u_{\alpha}(\hat{\underline{\underline{\theta}}}^{l},\underline{\theta}_{\alpha}^{'})
, \ \
\alpha\in\Phi_{l}^{(\hat{B},\hat{\underline{\underline{\theta}}}^{l})}
\} \nonumber
\end{equation}

\noindent{\bf Remark 2} a) In Step 1 of our policy we have to
compute the values of the objective function for finite number of
basic feasible solutions. These computations are not complicated
because the LP solution only needs the mean values of the
populations at this block and the randomization frequencies which
are as we know constants and depend only on which populations we
have in the BFS. We recall that if we have a BFS $b$, where $b=\{i,j
\} $ or $\{i\}$ then the optimal solution is $z^b=\mu_i x_i+\mu_j
x_j$ or $z^b=\mu_i$. Thus, in order to compute the value of the
objective function it is not required to solve the LPs but only to
compute and compare
 the corresponding $z^b$, using these explicit formulas.

The main result of this paper is that under the following conditions policy $\pi^0$ is asymptotically optimal.

To state condition C1  we need the definition of the index $J_{\alpha}(\underline{\underline{\theta}},\epsilon)$, of population $\alpha$. For any $\underline{\underline{\theta}} \in \Theta$,
 $ \epsilon>0$, an optimal matrix $B$ under $\underline{\underline{\theta}}$,
 and a
 $\rho(\underline{\underline{\theta}},\alpha,B)$, as in Lemma 1, we define:
  $\Theta_{\alpha}^{'}(\epsilon)=\{\underline{\theta}_{\alpha}^{'} 
: \,
\mu_{\alpha}^*(\underline{\underline{\theta}})-\epsilon <
\mu_{\alpha}(\underline{\theta}_{\alpha}^{'}) <
\mu_{\alpha}^*(\underline{\underline{\theta}}) +
\rho(\underline{\underline{\theta}},\alpha,B)-\epsilon \}$ and
\begin{equation}
J_{\alpha}(\underline{\underline{\theta}},\epsilon)=
\inf_{\underline{\theta}_{\alpha}^{'}\in \Theta_{\alpha}^{'}(\epsilon)}\{
I(\underline{\theta}_{\alpha},\underline{\theta}_{\alpha}^{'}):
z(\underline{\theta}_{\alpha}^{'})>z^*
(\underline{\underline{\theta}})-\epsilon \}.\nonumber
\end{equation}

From the definition of index
$J_{\alpha}(\hat{\underline{\underline{\theta}}}^{l},\epsilon)$,
where
$\alpha\in\Phi_{l}^{(\hat{B},\hat{\underline{\underline{\theta}}}^{l})}$,

\begin{equation}
J_{\alpha}(\hat{\underline{\underline{\theta}}}^{l},\epsilon)=
\inf_{\underline{\theta}_{\alpha}^{'}}\{
I(\hat{\underline{\theta}}_{\alpha}^{l},\underline{\theta}_{\alpha}^{'}):
z^{b_{\alpha}(\hat{\underline{\underline{\theta}}}^{l},\underline{\theta}_{\alpha}^{'})}>z^*
(\underline{\underline{\theta}})-\epsilon \}, \nonumber
\end{equation}
we have that
$u_{\alpha}(\hat{\underline{\underline{\theta}}}^{l},\underline{\theta}_{\alpha}^{'})>z^*
(\underline{\underline{\theta}})-\epsilon$ if and only if
$J_{\alpha}(\hat{\underline{\underline{\theta}}}^{l},\epsilon)< {\log
S_\pi(l-1)}/{T_\pi^{\alpha}(S_\pi(l-1))}$.\\

\noindent{\bf (C1) }
$\forall \ \underline{\underline{\theta}} \in
\Theta, \  i \notin s(\underline{\underline{\theta}})$  such
that $ \Delta\Theta_{i}(\underline{\underline{\theta}})=\emptyset,$
if
$\mu_{i}^*(\underline{\underline{\theta}})-\epsilon <
\mu_{i}(\underline{\theta}_{i}^{'}) <
\mu_{i}^*(\underline{\underline{\theta}}) +
\rho(\underline{\underline{\theta}},i,B)-\epsilon, $ \ $\forall \
\epsilon>0$, for some  $\underline{\theta}_{i}^{'} \in \Theta_{i}$, the following relation holds: $$ \lim_{\epsilon\rightarrow 0}
J_{i}(\underline{\underline{\theta}},\epsilon)=\infty.$$

\noindent{\bf (C2) } $  \forall i,$  $\forall \ \underline{\theta}_{i} \in \Theta_{i},
$ $\forall \ \epsilon > 0,$
$$P_{\underline{\theta}_i}(|\hat{\underline{\theta}}_{i}^{t}-\underline{\theta}_i|>\epsilon)=o(1/t), \mbox{ as } t\rightarrow \infty .$$

\noindent{\bf (C3) } $\forall$ $b_{\alpha}\in
 s(\underline{\underline{\theta}})$, $\forall i,$ $\forall \ \underline{\theta}_{i}
\in \Theta_{i}, $ $\forall \ \epsilon > 0,$ as $ t\rightarrow \infty
$
$$P_{\underline{\underline{\theta}}} (
z^{b_{\alpha}(\hat{\underline{\underline{\theta}}}^{j},\underline{\theta}_{\alpha}^{'})} \leq
z^* (\underline{\underline{\theta}})-\epsilon,
 \text{ for some }
j\leq t)=o(1/t).$$

 Next, we state and  prove the main theorem of the paper.\\

\noindent{\bf Theorem 2.}
Under conditions (C1),(C2), and (C3), and  policy $\pi^0$, defined above,  the following holds.
\begin{equation}
\limsup_{n\rightarrow\infty}\frac{R_{\pi^0}(\underline{\underline{\theta}},n)}{\log
n}\leq M(\underline{\underline{\theta}}), \text{ for all }
\underline{\underline{\theta}}\in\Theta.\nonumber
\end{equation}

\noindent{\bf Proof}

To establish the above inequality it is sufficient to show that for  policy $\pi^0$ the inequalities below hold.
\begin{equation}
\limsup_{n\rightarrow \infty}
\frac{E_{\underline{\underline{\theta}}}T^{j}_{\pi^0}(n)}{\log
n}\leq\frac{1}{K_{j}(\underline{\underline{\theta}})},
\ \forall  j\in D(\underline{\underline{\theta}}),\label{as4}
\end{equation}
\begin{equation}
\limsup_{n\rightarrow
\infty}\frac{E_{\underline{\underline{\theta}}}T^{j}_{\pi^0}(n)}{\log
n}=0, \ \forall  j\notin D(\underline{\underline{\theta}}), \label{as5}
\end{equation}
\begin{equation}
n c^0- E_{\underline{\underline{\theta}}}C_{\pi^0}(n)=o(\log
n).\label{as6}
\end{equation}
The proof of these inequalities is given in the appendix.

\noindent{\bf Remark 3} According to Remark $4b$ in \cite{bkmab96}
 condition (C2) is equivalent to C2' below which is easier to verify.

\noindent{\bf (C2')}
  $\forall \ \delta>0,$  as  $t\rightarrow \infty$
$$
 \sum_{j=1}^{t-1} P_{\underline{\theta}_i}
(b(\hat{\underline{\underline{\theta}}}^{j})\in
s(\underline{\underline{\theta}}),J_{i}(\hat{\underline{\underline{\theta}}}^{j},\epsilon)\leq
J_{i}(\underline{\underline{\theta}},\epsilon)-\delta) = o(\log t) .
$$

\noindent{\bf \large 6. Normal Distributions with known variances}

Assume the observations $X_{\alpha}^{j}$ from population
$\alpha$ are normally distributed with unknown means
$EX_{\alpha}^{j}=\theta_{\alpha}$ and known variances
$\sigma_{\alpha}^{2}$, i.e.,
$\underline{\theta}_{\alpha}=\theta_{\alpha}$,
$\mu_{\alpha}(\underline{\theta}_{\alpha})=\theta_{\alpha}$, and
$\Theta_{\alpha}=(-\infty,+\infty)$.
Given history $h_{l}$, define
\begin{equation}
\mu_{\alpha}(\hat{\theta}_{\alpha}^{l})=
\frac{\sum_{j=1}^{T_{\pi^0}^{\alpha}(S_{\pi^0}(l-1))}
X_{\alpha}^{j}}{T_{\pi^0}^{\alpha}(S_{\pi^0}(l-1))}.\nonumber
\end{equation}

Now from the definition of $\Theta_{\alpha}$, it follows that
$\Delta\Theta_{\alpha}(\underline{\theta})=(\theta_{\alpha}+\phi_{\alpha}^{B}(\underline{\theta}),
\theta_{\alpha}+\phi_{\alpha}^{B}(\underline{\theta}) +
\rho(\underline{\theta},\alpha,B))$ for any optimal matrix $B$ under
$\underline{\theta}$, therefore $D(\underline{\theta})=\{ 1,...,k
\}$, $\forall$ $\underline{\theta}\in\Theta$. Thus, we can see from
the structure of the sets $\Theta_{\alpha}$ and
$\Delta\Theta_{\alpha}(\underline{\theta})$ that condition (C1)
is satisfied.

Also, we have:
$$I(\theta_{\alpha},\theta_{\alpha}^{'})
=\frac{(\theta_{\alpha}^{'}-\theta_{\alpha})^2}{2\sigma_{\alpha}^{2}}$$
$$K_{\alpha}(\underline{\theta})=
\frac{(\phi_{\alpha}^{B}(\underline{\theta}))^2}{2\sigma_{\alpha}^{2}}.$$
Therefore our indices are equal to
$$
u_{\alpha}(\hat{\underline{\theta}}^{l},\theta_{\alpha}^{K_{\alpha}})=
z^{b_{\alpha}(\hat{\underline{\theta}}^{l},\theta_{\alpha}^{K_{\alpha}})},$$
 where
$$ \theta_{\alpha}^{K_{\alpha}}=\hat{\theta}_{\alpha}^{l}+
\sigma_{\alpha}\left(\frac{2\log
S_{\pi^0}(l-1)}{T_{\pi^0}^{\alpha}(S_{\pi^0}(l-1))}\right)^{1/2}.$$
For example, if
$b_{\alpha}(\hat{\underline{\theta}}^{l},\theta_{\alpha}^{K_{\alpha}})=\{\alpha,j\}$,
then
$z^{b_{\alpha}(\hat{\underline{\theta}}^{l},\theta_{\alpha}^{K_{\alpha}})}=
\theta_{\alpha}^{K_{\alpha}} x_{\alpha} + \hat{\theta}_{j}^{l} x_j$
and $z^{*}(\underline{\theta})=\theta_{\alpha}x_{\alpha}+\theta_j
x_j$. Therefore for
$b_{\alpha}(\hat{\underline{\theta}}^{l},\theta_{\alpha}^{K_{\alpha}})\in
s(\underline{\theta})$ and from the structure of
$z^{b_{\alpha}(\hat{\underline{\theta}}^{l},\theta_{\alpha}^{K_{\alpha}})}$
the index is a sum of normal distributions which is also normal or a
normal distribution and from the tail of normal distribution
condition (C3) is satisfied.

According to Remark 3 the next sum of probabilities is equivalent to
condition (C2)
\begin{eqnarray}
&&\sum_{t=2}^{L_n} P_{\theta_i} (b(\hat{\underline{\theta}}^{t})\in
s(\underline{\theta}),J_{i}(\hat{\underline{\theta}}^{t},\epsilon)\leq
J_{i}(\underline{\theta},\epsilon)-\delta)\nonumber\\
&& =\sum_{t=2}^{L_n}P_{\theta_i} (
b(\hat{\underline{\theta}}^{t})\in s(\underline{\theta}),
|\hat{\theta}_{i}^{t}-\theta_i|>\xi ),\xi>0,\nonumber
\end{eqnarray}
where the equality follows after some algebra because of the normal
distribution and the explicit form of $I(\hat{\theta}_{i}^{t},\theta_{i}^{'})$ in this case:
\begin{eqnarray}
&&J_{i}(\hat{\underline{\theta}}^{t},\epsilon)=
\inf_{\theta_{i}^{'}}\{ I(\hat{\theta}_{i}^{t},\theta_{i}^{'}):
z^{b_{i}(\hat{\underline{\theta}}^{t},\theta_{i}^{'})}>z^*
(\underline{\theta})-\epsilon \} \leq \nonumber\\
&&J_{i}(\underline{\theta},\epsilon)=\inf_{\theta_{i}^{'}}\{
I(\theta_{i},\theta_{i}^{'}):
z^{b_{i}(\underline{\theta},\theta_{i}^{'})}>z^*
(\underline{\theta})-\epsilon \}-\delta.\nonumber
\end{eqnarray}
Also, we have that $\hat{\theta}_{i}^{t}$ is the average of iid
random normal variables with mean $\theta_i$ thus
\begin{eqnarray}
P_{\theta_i}^{\pi^0}(|\hat{\theta}_{i}^{t}-\theta_i|>\xi)& \leq&
P_{\theta_i}^{\pi^0}(|\hat{\theta}_{i}^{l}-\theta_i|>\xi,
\text{ for some }l\leq t)\nonumber\\
&\leq&\sum_{l=1}^{t}P_{\theta_i}^{\pi^0}(|\hat{\theta}_{i}^{l}-\theta_i|>\xi)=o(1/t),\nonumber
\end{eqnarray}
where the last equality follows from is a consequence of the tail
inequality $1 - \Phi(x) < \Phi(x)/x$ for the standard normal
distribution. Thus, we can see that condition (C2) holds.\\

\noindent
{\bf Summary of Policy} At the beginning we take an ISB block. Then
at the beginning of block $l$ we take
$$ z^{b(\hat{\underline{\underline{\theta}}}^{l})}=
\max_{\widetilde{b}_i (\hat{\underline{\underline{\theta}}}^{l})}\{
z^{\widetilde{b}_i (\hat{\underline{\underline{\theta}}}^{l})}: \
\widetilde{T}_\pi^{\widetilde{b}_i
(\hat{\underline{\underline{\theta}}}^{l})}(l)\geq \tau(l-1) \}$$
and find our indices
$$ u_{\alpha}(\hat{\underline{\theta}}^{l},\theta_{\alpha}^{K_{\alpha}})=
z^{b_{\alpha}(\hat{\underline{\theta}}^{l},\theta_{\alpha}^{K_{\alpha}})},$$
where
\begin{equation}\label{thetaka} \theta_{\alpha}^{K_{\alpha}}=\hat{\theta}_{\alpha}^{l}+
\sigma_{\alpha}\left(\frac{2\log
S_{\pi^0}(l-1)}{T_{\pi^0}^{\alpha}(S_{\pi^0}(l-1))}\right)^{1/2}.
\end{equation}
Finally, we choose to employ as block $l$ the
$\argmax_{\alpha} \{
u_{\alpha}(\hat{\underline{\underline{\theta}}}^{l},\underline{\theta}_{\alpha}^{K_{\alpha}})\}$. \\

\noindent{\bf Remark 4} In the case in which   $\sigma_{\alpha}$ are
unknown, we expect that a (log - rate regret) f-UF policy can be
obtained by replacing
 $\sigma_{\alpha}$ in Eq. {\ref{thetaka}) by a constant times
  $\hat{\sigma}_{\alpha},$ as in \cc{Auer02b}. This work is currently in progress.

\newpage

\bibliography{mab2015}
\newpage
\noindent{\bf \large Appendix: Proofs}

{\bf Lemma 1} For any $\underline{\underline{\theta}}$, and any
optimal matrix $B$ under $\underline{\underline{\theta}}$, $\exists$
$\rho=\rho(\underline{\underline{\theta}},\alpha,B)>0$ such that for
any
$\underline{\theta}_{\alpha}^{'}\in\Delta\Theta_{\alpha}(\underline{\underline{\theta}}):$\\
$\textbf{(i)}$ $\phi^B_{j}
(\underline{\underline{\theta}}^{'})=\phi^B_{j}
(\underline{\underline{\theta}})\geq 0, \ \forall \ j\neq\alpha $
and
$\phi^B_{\alpha} (\underline{\underline{\theta}}^{'})=\phi^B_{\alpha} (\underline{\underline{\theta}})
+\mu_{\alpha}(\underline{\theta}_{\alpha})-\mu_{\alpha}(\underline{\theta}_{\alpha}^{'})<0$,\\
$\textbf{(ii)}$
$\mu_{\alpha}^*(\underline{\underline{\theta}})<\mu_{\alpha}(\underline{\theta}_{\alpha}^{'})<\mu_{\alpha}^*(\underline{\underline{\theta}})+\rho$,
where
$\mu_{\alpha}^*(\underline{\underline{\theta}})=\phi^B_{\alpha}
(\underline{\underline{\theta}})
+\mu_{\alpha}(\underline{\theta}_{\alpha})$.

{\bf Proof} $(i)$ It is obvious that $\phi^B_{j}
(\underline{\underline{\theta}}^{'})=\phi^B_{j}
(\underline{\underline{\theta}})\geq 0, \ \forall \ j\neq\alpha$
because we only change the parameter of population $\alpha$ and
$\phi^B_{j}
(\underline{\underline{\theta}}^{'})=\phi^B_{j}(\underline{\underline{\theta}})\equiv
c^{j} \lambda^B + g^B - \mu_{j}(\underline{\theta}_{j})$.

For a population $\alpha\in B(\underline{\underline{\theta}})$ we
have that $\alpha\notin b$, for any $b\in
s(\underline{\underline{\theta}})$. Therefore $\phi^B_{\alpha}
(\underline{\underline{\theta}})\equiv c^{\alpha} \lambda^B + g^B -
\mu_{\alpha}(\underline{\theta}_{\alpha})
> 0$, for any $B$ corresponding to $b$.

Now, any optimal $b\in s(\underline{\underline{\theta}})$ is not
optimal under
$\underline{\underline{\theta}}^{'}=(\underline{\theta}_1,
\ldots,\underline{\theta}_{\alpha}^{'},\ldots,\underline{\theta}_k)$,
for any $\underline{\theta}_{\alpha}^{'}\in
\Delta\Theta_{\alpha}(\underline{\underline{\theta}})$, thus
$s(\underline{\underline{\theta}}^{'})=\{ b^{'} \}$ where
$b^{'}\notin s(\underline{\underline{\theta}})$.

Therefore, for any optimal matrix $B$ under
$\underline{\underline{\theta}}$ we have that $\phi^B_{\alpha}
(\underline{\underline{\theta}}^{'})\equiv c^{\alpha} \lambda^B +
g^B - \mu_{\alpha}(\underline{\theta}_{\alpha}^{'})< 0$ because $B$
is not optimal under $\underline{\underline{\theta}}^{'}$.

Now from $\phi^B_{\alpha} (\underline{\underline{\theta}})=
c^{\alpha} \lambda^B + g^B -
\mu_{\alpha}(\underline{\theta}_{\alpha})$ we have that
$\phi^B_{\alpha}
(\underline{\underline{\theta}}^{'})=\phi^B_{\alpha}
(\underline{\underline{\theta}})
+\mu_{\alpha}(\underline{\theta}_{\alpha})-\mu_{\alpha}(\underline{\theta}_{\alpha}^{'})
< 0$.

$(ii)$ Consider first the case where   $b=\{ i,j \}$ is an optimal
solution under $\underline{\underline{\theta}}$ with corresponding
optimal matrix $B=B(\underline{\underline{\theta}})$, and $b'=\{
i,\alpha \}$ is an optimal solution under
$\underline{\underline{\theta}}^{'}$ with corresponding optimal
matrix $B'=B(\underline{\underline{\theta}}')$. From $(i)$ we have
that $z^* (\underline{\underline{\theta}}^{'})>z^*
(\underline{\underline{\theta}})$ iff
$\mu_{\alpha}(\underline{\theta}_{\alpha}^{'})>\mu_{\alpha}^*(\underline{\underline{\theta}})$.

Since $b'$ is uniquely optimal under
$\underline{\underline{\theta}}^{'}$ we have that
$\phi_{s}^{B'}(\underline{\underline{\theta}}^{'})>0$, for any
$s\neq i,\alpha$. Now in order for that condition to hold we use
that $\phi_{s}^{B}(\underline{\underline{\theta}})>0$ for any $s\neq
i,j$ and we have that for $s>i$ it suffices that
$\mu_{\alpha}^*(\underline{\underline{\theta}})<\mu_{\alpha}(\underline{\theta}_{\alpha}^{'})$,
while for $s<i$ we must have
$\mu_{\alpha}^*(\underline{\underline{\theta}})<\mu_{\alpha}(\underline{\theta}_{\alpha}^{'})<\mu_{\alpha}^*(\underline{\underline{\theta}})+\rho$,
where $\rho$ is a positive constant. Thus, if
$\mu_{\alpha}^*(\underline{\underline{\theta}})<\mu_{\alpha}(\underline{\theta}_{\alpha}^{'})<\mu_{\alpha}^*(\underline{\underline{\theta}})+\rho$
then $\phi_{s}^{B'}(\underline{\underline{\theta}}^{'})>0$ for any
$s$.

The other cases where the population $\alpha$ is a population with
cost lower than $c^0$ and the optimal solution under
$\underline{\underline{\theta}}^{'}$ has this form $b'=\{ \alpha, j
\}$ or $b'=\{ \alpha \}$ follows with the same arguments as in the
previous paragraph.\\
$\square$

{\bf Lemma 2} Assume $b$ is uniquely optimal BFS and $B$ any optimal
matrix under $\underline{\underline{\theta}}$. Then\\
\emph{\textbf{(i)}} if $B=\left(
\begin{array}{cc} c^i & c^j\\ 1 &
1\\
\end{array} \right)$, for some $i\leq d < j$ then $\lambda^B >0$,\\
\emph{\textbf{(ii)}} if $B=\left( \begin{array}{cc} c^i & 1\\ 1 & 0
\\
\end{array} \right)$, for some $i\leq d$ then $\lambda^B =0$.\label{lemma2}

{\bf Proof} \emph{\textbf{(i)}} Let
$\underline{\underline{\theta}}:$
$s(\underline{\underline{\theta}})=\{ b \}$, $b=(i,j)$ for $i\leq d
< j$, then $\lambda^B >0$ because if $\lambda^B =0$ we must have
more
than one solutions in the primal, which cannot occur because $b$ is uniquely optimal.\\
\emph{\textbf{(ii)}} Let $\underline{\underline{\theta}}:$
$s(\underline{\underline{\theta}})=\{ b \}$, $b=(i)$ for $i\leq d$,
then $\lambda^B =0$ from the dual solution and
$\phi_{j}^{B}(\underline{\underline{\theta}})>0$ for all $j\neq
i$.\\
$\square$

We recall for the next Proposition
\begin{eqnarray}
z^{*}(\underline{\underline{\theta}}) & = & \max
\sum_{j=1}^{k}\mu_{j}(\underline{\theta}_j) x_j \nonumber\\
&&\sum_{j=1}^{k}c^j x_j + y = c^0 \nonumber\\
&&\sum_{j=1}^{k}x_j =1 \nonumber\\
&&x_j \geq 0,\forall j, \ y\geq 0, \nonumber
\end{eqnarray}
and that a necessary and sufficient condition for a uniformly fast
policy $\pi$ is that for $\underline{\underline{\theta}}\in\Theta$
and any optimal BFS $b$ under $\underline{\underline{\theta}}$,
\begin{equation}
\phi_{j}^{B}(\underline{\underline{\theta}})\lim_{n\rightarrow\infty}\frac{E_{\underline{\underline{\theta}}}T_{\pi}^{j}(n)}{n^a}=0,
\text{ for all }  a>0, \ j\notin b,\label{ug1}
\end{equation}
and also,
\begin{equation}
\lambda^{B}\lim_{n\rightarrow\infty}\frac{\sum_{j\in b}(c^0 -
c^j)E_{\underline{\underline{\theta}}}T_{\pi}^{j}(n)}{n^a}=0, \text{
for all }B \text{ corresponding to }b.\label{ug2}
\end{equation}

{\bf Proposition 1} For any uniformly fast policy $\pi$ and for all
$\underline{\underline{\theta}}\in\Theta$ we have that for
$\alpha\in D(\underline{\underline{\theta}})$, any
$\underline{\underline{\theta}}^{'}\in\Delta(\underline{\underline{\theta}})$
and for all positive $\beta_n=o(n)$ it is true that
\begin{equation}
P_{\underline{\underline{\theta}}^{'}}[T^{\alpha}_{\pi}(n)
<\beta_n]=o(n^{a-1}), \text{ for all }a>0. \nonumber
\end{equation}

{\bf Proof} Let $\alpha\in D(\underline{\underline{\theta}})$,
$\theta_{\alpha}^{'}\in\Delta\Theta_{\alpha}(\underline{\underline{\theta}})$.
Because of the definition of $\Delta\Theta_{\alpha}(\underline{\underline{\theta}})$
we must have a $b^{'}$ which is uniquely optimal under
$\underline{\underline{\theta}}^{'}$
($s(\underline{\underline{\theta}}^{'})=\{ b^{'} \}$) and $\alpha\in
b^{'}$. Then we have two cases for the uniquely optimal solution
$b^{'}$.

For the first case where $b^{'}=\{ \alpha \}$ if $b^{'}$ is
nondegenerate then the basic matrix $B^{'}=\left( \begin{array}{cc}
c^{\alpha} & 1\\ 1 & 0
\\
\end{array} \right)$ and from Lemma $2$ for a uniformly fast policy $\lambda^B=0$ thus,
\begin{equation}
E_{\underline{\underline{\theta}}^{'}}T^{j}_{\pi}(n)=o(n^{a}),
\text{ for all } a>0, \text{ for all } j\notin b^{'}.\nonumber
\end{equation}
If $b^{'}$ is degenerate then it must be true that $c^{\alpha}=c^0$
if we consider any matrix $B^{'}=\left( \begin{array}{cc} c^{\alpha}
& c^j\\ 1 &
1\\
\end{array} \right)$ then $\lambda_{B^{'}}>0$ thus
$(c^0-c^{j})E_{\underline{\underline{\theta}}^{'}}T^{j}_{\pi}(n)+
(c^0
-c^{\alpha})E_{\underline{\underline{\theta}}^{'}}T^{\alpha}_{\pi}(n)
=o(n^{a})$ and since $c^0=c^{\alpha}$ we have that
$E_{\underline{\underline{\theta}}^{'}}T^{j}_{\pi}(n)=o(n^{a})$. Moreover
from \eqref{ug1}
$E_{\underline{\underline{\theta}}^{'}}T^{i}_{\pi}(n)=o(n^{a})$, for
all $i\neq j,\alpha$, thus
$E_{\underline{\underline{\theta}}^{'}}T^{j}_{\pi}(n)=o(n^{a})$, for
all $j\neq \alpha$.

Therefore,
\begin{equation}
n-E_{\underline{\underline{\theta}}^{'}}T^{\alpha}_{\pi}(n)
=o(n^{a}), \text{ for all } a>0.\label{prop1}
\end{equation}

It is also true that
\begin{eqnarray}
E_{\underline{\underline{\theta}}^{'}}T^{\alpha}_{\pi}(n) &=&
\sum_{k=1}^{n}k\, P_{\underline{\underline{\theta}}^{'}}[T^{\alpha}_{\pi}(n) =k]\nonumber\\
&=&\sum_{k=1}^{\lfloor\beta_n\rfloor}k\,
P_{\underline{\underline{\theta}}^{'}}[T^{\alpha}_{\pi}(n)
=k]+\sum_{k=\lfloor\beta_n\rfloor+1}^{n}k\,
P_{\underline{\underline{\theta}}^{'}}[T^{\alpha}_{\pi}(n) =k] \nonumber\\
&\leq&\beta_n
P_{\underline{\underline{\theta}}^{'}}[T^{\alpha}_{\pi}(n) \leq
\beta_n]+
n P_{\underline{\underline{\theta}}^{'}}[T^{\alpha}_{\pi}(n) >\beta_n]\nonumber\\
&=&n-(n-\beta_n)P_{\underline{\underline{\theta}}^{'}}[T^{\alpha}_{\pi}(n)
\leq \beta_n].\nonumber
\end{eqnarray}
Therefore
\begin{equation}
n-E_{\underline{\underline{\theta}}^{'}}T^{\alpha}_{\pi}(n)\geq(n-\beta_n)P_{\underline{\underline{\theta}}^{'}}[T^{\alpha}_{\pi}(n)\leq
\beta_n].\label{prop2}
\end{equation}

From \eqref{prop1} and \eqref{prop2} we obtain
\begin{equation}
(n-\beta_n)P_{\underline{\underline{\theta}}^{'}}[T^{\alpha}_{\pi}(n)
\leq \beta_n]=o(n^a), \text{ for all }a>0, \nonumber
\end{equation}
thus
\begin{equation}
P_{\underline{\underline{\theta}}^{'}}[T^{\alpha}_{\pi}(n) \leq
\beta_n]=o(n^{a-1}), \text{ for all }a>0. \nonumber
\end{equation}

We next consider the case $b^{'}=\{ j_0,\alpha \}$ with $c^{j_0}<c^0<c^{\alpha}$. (The case $c^{\alpha}<c^0<c^{j_0}$ is completely analogous). We have from Lemma $2$ that for a uniformly fast
policy $\lambda^B>0$, thus

\begin{equation}
E_{\underline{\underline{\theta}}^{'}}T^{j}_{\pi}(n)=o(n^{a}), \
\forall \ a>0, \ \forall \ j\notin b^{'}=\{j_0,\alpha \}\label{lem1}
\end{equation}
and

\begin{equation}
(c^0
-c^{j_0})E_{\underline{\underline{\theta}}^{'}}T^{j_0}_{\pi}(n)+
(c^0
-c^{\alpha})E_{\underline{\underline{\theta}}^{'}}T^{\alpha}_{\pi}(n)
=o(n^{a}), \ \forall \ a>0.\label{lem2}
\end{equation}
If we sum \eqref{lem1} for all $j\neq\alpha,j_0$ it follows that

\begin{equation}
n-E_{\underline{\underline{\theta}}^{'}}T^{j_0}_{\pi}(n)-E_{\underline{\underline{\theta}}^{'}}T^{\alpha}_{\pi}(n)
=\varepsilon_n, \text{ where } \varepsilon_n=o(n^{a}), \ \forall \
a>0.\label{lem3}
\end{equation}
Dividing \eqref{lem2} with $c^{\alpha}-c^{j_0}$ and using
\eqref{lem3}, we obtain after some algebra the following two
equalities

\begin{eqnarray}
&nx_{j_0}^{'}-E_{\underline{\underline{\theta}}^{'}}T^{j_0}_{\pi}(n)=o(n^{a}),& \label{lem4}\\
&nx_{\alpha}^{'}-E_{\underline{\underline{\theta}}^{'}}T^{\alpha}_{\pi}(n)=o(n^{a}),
\ \forall \ a>0.&\nonumber
\end{eqnarray}
where $x_{j_0}^{'}=\frac{c^{\alpha}-c^0}{c^{\alpha}-c^{j_0}}$ and
$x_{\alpha}^{'}=\frac{c^0-c^{j_0}}{c^{\alpha}-c^{j_0}}$ are the
probabilities which correspond to optimal solution $b^{'}$ of linear
program \eqref{lp} under $\underline{\underline{\theta}}^{'}$.

For any $n$ let

\begin{equation}
\Gamma_{n}^{\pi}=\sum_{j\neq\alpha,j_0}T^{j}_{\pi}(n), \text{ and }
F_n^{\pi} =\sum_{j\neq\alpha,j_0}(c^0 -c^j)T^{j}_{\pi}(n).\nonumber
\end{equation}
Thus, it is obvious that

\begin{equation}
F_{n}^{\pi}\leq\Gamma_{n}^{\pi}(c^0-c^1).\nonumber
\end{equation}
Furthermore, from \eqref{lem3}

\begin{equation}
E_{\underline{\underline{\theta}}^{'}}\Gamma_{n}^{\pi}=o(n^{a}), \
\forall \ a>0.\label{lem5}
\end{equation}
Now, we know that

\begin{equation}
nc^0-C_{\pi}(n)=F_n^{\pi} +(c^0-c^{\alpha})T^{\alpha}_{\pi}(n)
+(c^0-c^{j_0})T^{j_0}_{\pi}(n),\nonumber
\end{equation}
and from $nc^0-C_{\pi}(n)\geq0, \ \forall \ n$, we have that

\begin{equation}
(c^{\alpha}-c^0)T^{\alpha}_{\pi}(n) \leq F_n^{\pi}
+(c^0-c^{j_0})T^{j_0}_{\pi}(n),\nonumber
\end{equation}
therefore

\begin{eqnarray}
\frac{c^{\alpha}-c^0}{c^{\alpha}-c^{j_0}}
T^{\alpha}_{\pi}(n)  &\leq& \frac{F_n^{\pi}}{c^{\alpha}-c^{j_0}}+ \frac{c^0-c^{j_0}}{c^{\alpha}-c^{j_0}}T^{j_0}_{\pi}(n) \nonumber\\
x_{j_0}^{'}T^{\alpha}_{\pi}(n) &\leq&\frac{F_n^{\pi}}{c^{\alpha}-c^{j_0}}+ x_{\alpha}^{'}T^{j_0}_{\pi}(n)\nonumber \\
(1-x_{\alpha}^{'})T^{\alpha}_{\pi}(n) &\leq&\frac{F_n^{\pi}}{c^{\alpha}-c^{j_0}}+ x_{\alpha}^{'}T^{j_0}_{\pi}(n)\nonumber \\
T^{\alpha}_{\pi}(n) &\leq&\frac{F_n^{\pi}}{c^{\alpha}-c^{j_0}}+ x_{\alpha}^{'}(T^{\alpha}_{\pi}(n) +T^{j_0}_{\pi}(n))\nonumber \\
T^{\alpha}_{\pi}(n) &\leq&\frac{F_n^{\pi}}{c^{\alpha}-c^{j_0}}+ x_{\alpha}^{'}(n- \Gamma_n^{\pi})\nonumber \\
T^{\alpha}_{\pi}(n) &\leq&n
x_{\alpha}^{'}+\frac{F_n^{\pi}}{c^{\alpha}-c^{j_0}}-
x_{\alpha}^{'}\Gamma_n^{\pi}.\nonumber
\end{eqnarray}
Recall $F_n^{\pi}\leq\Gamma_n^{\pi}(c^0-c^1)$, thus

\begin{eqnarray}
T^{\alpha}_{\pi}(n) &\leq&n x_{\alpha}^{'}+
\frac{\Gamma_n^{\pi}(c^0-c^1)}{c^{\alpha}-c^{j_0}}- x_{\alpha}^{'}\Gamma_n\nonumber\\
T^{\alpha}_{\pi}(n) &\leq&n x_{\alpha}^{'} +\Gamma_n^{\pi}
\rho(j_0,\alpha)\nonumber
\end{eqnarray}
where $\rho(j_0,\alpha)=\frac{c^{j_0}-c^1}{c^{\alpha}-c^{j_0}}\geq
0$.

Finally,
\begin{equation}
nx_{\alpha}^{'}-T^{\alpha}_{\pi}(n) +\Gamma_{n}^{\pi}
\rho(j_0,\alpha)\geq 0. \label{lem6}
\end{equation}
Thus, from Markov inequality, for any positive $\beta_n=o(n)$

\begin{eqnarray}
&P_{\underline{\underline{\theta}}^{'}}(n
x_{\alpha}^{'}-T^{\alpha}_{\pi}(n) +\Gamma_{n}^{\pi}
\rho(j_0,\alpha)
\geq n x_{\alpha}^{'}-\beta_n)&\nonumber\\
&\leq \frac{E_{\underline{\underline{\theta}}^{'}}(n x_{\alpha}^{'}-
T^{\alpha}_{\pi}(n) +\Gamma_{n}^{\pi} \rho(j_0,\alpha))}{n x_{\alpha}^{'}-\beta_n}&\nonumber\\
&=\frac{o(n^{a})}{n x_{\alpha}^{'}-\beta_n}=o(n^{a-1}), \ \forall \
a>0.&\nonumber
\end{eqnarray}
Therefore

\begin{equation}
P_{\underline{\underline{\theta}}^{'}}(T^{\alpha}_{\pi}(n) \leq
\beta_n)\leq
P_{\underline{\underline{\theta}}^{'}}(T^{\alpha}_{\pi}(n) \leq
\beta_n + \Gamma_{n}^{\pi} \rho(j_0,\alpha)) = o(n^{a-1}), \forall \
a>0.\nonumber
\end{equation}
Substituting $T^{\alpha}_{\pi}(n)
=n-\Gamma_{n}^{\pi}-T^{j_0}_{\pi}(n)$ into \eqref{lem6} we have

\begin{equation}
T^{j_0}_{\pi}(n)-n
x_{j_0}^{'}+(1+\rho(j_0,\alpha))\Gamma_{n}^{\pi}\geq 0,\nonumber
\end{equation}
then

\begin{equation}
P_{\underline{\underline{\theta}}^{'}}(T^{j_0}_{\pi}(n)\leq
\beta_n)=P_{\underline{\underline{\theta}}^{'}}(Z_n^{\pi}\leq
\beta_n -n x_{j_0}^{'}+(1+\rho(j_0,\alpha))\Gamma_{n}^{\pi})
,\nonumber
\end{equation}

where

\begin{equation}
Z_n^{\pi}=T^{j_0}_{\pi}(n)-n
x_{j_0}^{'}+(1+\rho(j_0,\alpha))\Gamma_{n}^{\pi}\geq 0,\nonumber
\end{equation}
and

\begin{equation}
E_{\underline{\underline{\theta}}^{'}}Z_n^{\pi}=o(n^{a}), \ \forall
\ a>0 \text{ from } \eqref{lem4} \text{ and } \eqref{lem5}.\nonumber
\end{equation}
Let,

\begin{equation}
V_n^{\pi}=\{ Z_n^{\pi}\leq \beta_n -n
x_{j_0}^{'}+(1+\rho(j_0,\alpha))\Gamma_{n}^{\pi} \}, \text{ then
}\nonumber
\end{equation}

\begin{eqnarray}
P_{\underline{\underline{\theta}}^{'}}(V_n^{\pi})&=&P_{\underline{\underline{\theta}}^{'}}
(V_n^{\pi} \cap \{ \Gamma_{n}^{\pi}\leq n \delta
\})+P_{\underline{\underline{\theta}}^{'}}
(V_n^{\pi} \cap \{ \Gamma_{n}^{\pi}> n \delta \})\nonumber\\
&\leq&P_{\underline{\underline{\theta}}^{'}}(V_n^{\pi} \cap \{
\Gamma_{n}^{\pi}\leq n \delta
\})+P_{\underline{\underline{\theta}}^{'}}(\Gamma_{n}^{\pi}> n
\delta)\label{lem7}
\end{eqnarray}
where $0<\delta<\frac{x_{j_0}^{'}}{1+\rho(j_0,\alpha)}$ and using
\eqref{lem5} we have that

\begin{eqnarray}
P_{\underline{\underline{\theta}}^{'}}(\Gamma_{n}^{\pi}> n
\delta)&\leq&
 \frac{E_{\underline{\underline{\theta}}^{'}}\Gamma_{n}^{\pi}}{n\delta}\nonumber\\
&=&\frac{o(n^{a})}{n\delta}=o(n^{a-1}), \ \forall \ a>0.\label{lem8}
\end{eqnarray}
Let,

\begin{eqnarray}
G_n^{\pi}&=&\{ V_n^{\pi} \cap \{ \Gamma_{n}^{\pi}\leq n \delta \} \}\nonumber\\
&=&\{ Z_n^{\pi}\leq \beta_n -n
x_{j_0}^{'}+(1+\rho(j_0,\alpha))\Gamma_{n}^{\pi}
 \text{ and }\Gamma_{n}^{\pi}\leq n \delta \}\nonumber\\
&\subseteq&\{ Z_n^{\pi}\leq \beta_n+[(1+\rho(j_0,\alpha))\delta-x_{j_0}^{'}]n \},\nonumber\\
&=&\{ Z_n^{\pi} \leq\beta_n-\varphi n \},\nonumber
\end{eqnarray}
where
\begin{equation}
\varphi=x_{j_0}^{'}-(1+\rho(j_0,\alpha))\delta
>x_{j_0}^{'}-(1+\rho(j_0,\alpha))\frac{x_{j_0}^{'}}{1+\rho(j_0,\alpha)}=0.\nonumber
\end{equation}
Now for any positive $\beta_n=o(n)$,

\begin{equation}
\exists \ n_0: \ \beta_n-n\varphi<0, \ \forall \ n>n_0\nonumber
\end{equation}
and we have that

\begin{equation}
P_{\underline{\underline{\theta}}^{'}}(G_n^{\pi})=0,  \forall \
n>n_0(\varphi),\nonumber
\end{equation}
thus from \eqref{lem7},\eqref{lem8}

\begin{equation}
P_{\underline{\underline{\theta}}^{'}}(V_n^{\pi})\leq o(n^{a-1}), \
\forall \ a>0.\nonumber
\end{equation}
Finally,

\begin{equation}
P_{\underline{\underline{\theta}}^{'}}(T^{j_0}_{\pi}(n)\leq
\beta_n)=o(n^{a-1}), \ \forall \ a>0, \text{ for any positive }
\beta_n=o(n).\nonumber
\end{equation}
$\square$

{\bf Lemma 3} If
$P_{\underline{\underline{\theta}}^{'}}[T^{\alpha}_{\pi}(n)
<\beta_n]=o(n^{a-1}),$ for all $a>0$ and positive $\beta_n=o(n)$,
then
\begin{equation}
\lim_{n\rightarrow\infty}P_{\underline{\underline{\theta}}}[T^{\alpha}_{\pi}(n)<\frac{\log
n}{K_{\alpha}(\underline{\underline{\theta}})}]=0,\nonumber
\end{equation}
for all $\underline{\underline{\theta}}\in\Theta$ and
$\alpha\in\Delta(\underline{\underline{\theta}})$.

{\bf Proof} If we take $\beta_n=\frac{\log
n}{K_{\alpha}(\underline{\underline{\theta}})}$, then
$P_{\underline{\underline{\theta}}^{'}}[T^{\alpha}_{\pi}(n)
<\frac{\log
n}{K_{\alpha}(\underline{\underline{\theta}})}]=o(n^{a-1})$ and
using a change of measure from $\underline{\underline{\theta}}^{'}$
to $\underline{\underline{\theta}}$ and following the arguments in
  \cite{bkmab96,Lai85}  we have that
\begin{equation}
\lim_{n\rightarrow\infty}P_{\underline{\underline{\theta}}}[T^{\alpha}_{\pi}(n)<\frac{\log
n}{K_{\alpha}(\underline{\underline{\theta}})}] =0.\nonumber
\end{equation}
$\square$

We recall for Theorem 2 that
\begin{eqnarray}
&1.&\limsup_{n\rightarrow \infty}
\frac{E_{\underline{\underline{\theta}}}T^{j}_{\pi}(n)}{\log
n}\leq\frac{1}{K_{j}(\underline{\underline{\theta}})},
\text{ for all } j\in D(\underline{\underline{\theta}}),\label{as4}\\
&2.&\limsup_{n\rightarrow
\infty}\frac{E_{\underline{\underline{\theta}}}T^{j}_{\pi}(n)}{\log
n}=0,
\text{ for all } j\notin D(\underline{\underline{\theta}}), \label{as5}\\
&3.&n c^0- E_{\underline{\underline{\theta}}}C_{\pi}(n)=o(\log
n).\label{as6}
\end{eqnarray}
From the definition of $T^{\alpha}_{\pi}(n) $ we can see that
\begin{equation}
T_\pi^\alpha(S_\pi(L_n)) \leq T^{\alpha}_{\pi}(n)  \leq
T_\pi^\alpha(S_\pi(L_n))+M_{\alpha},\label{as3}
\end{equation}
where $M_{\alpha}$ is the maximum number of times where population
$\alpha$ appears in every block.

We have derived $\widetilde{T}_\pi^b(L_n)$ as:
\begin{eqnarray}
\widetilde{T}_\pi^b(L_n)&=&\sum_{t=2}^{L_n}1\{ \pi_{t}^{0}= b,
b(\hat{\underline{\underline{\theta}}}^{t})\notin
s(\underline{\underline{\theta}}) \}
+ \sum_{t=2}^{L_n}1\{ \pi_{t}^{0}=b, b(\hat{\underline{\underline{\theta}}}^{t})\in s(\underline{\underline{\theta}}) \}\nonumber \\
&\leq&\sum_{t=2}^{L_n}1\{
b(\hat{\underline{\underline{\theta}}}^{t})\notin
s(\underline{\underline{\theta}}) \} + \sum_{t=2}^{L_n}1\{
\pi_{t}^{0}=b, b(\hat{\underline{\underline{\theta}}}^{t})\in
s(\underline{\underline{\theta}}) \}.\label{as7}
\end{eqnarray}
Finally, a policy $\pi$ is called feasible if
\begin{equation}
\label{eq:feasible}
 \frac{C_{\pi}(n)}{n} \leq c^0, \ \forall \ n=1,2,\ldots.
\end{equation}

{\bf Theorem 2}\label{thm2}
Under conditions (C1),(C2), and (C3), policy $\pi^0$ satisfies:\\
\begin{equation}
\limsup_{n\rightarrow\infty}\frac{R_{\pi^0}(\underline{\underline{\theta}},n)}{\log
n}\leq M(\underline{\underline{\theta}}), \text{ for all }
\underline{\underline{\theta}}\in\Theta.\nonumber
\end{equation}

{\bf Proof} We need to prove \eqref{as4}, \eqref{as5} and
\eqref{as6}. From the \eqref{as3}, \eqref{as7} and Lemmas $4$ and
$5$ we have proved the relations  \eqref{as4} and \eqref{as5}.
Equation \eqref{as6} follows from \eqref{eq:feasible} the
feasibility of $\pi^0$ and block policies.\\
$\square$

{\bf Lemma 4}
Under conditions (C1),(C2), policy $\pi^0$ satisfies:\\
\begin{eqnarray}
&&\limsup_{n\rightarrow
\infty}\frac{E_{\underline{\underline{\theta}}}\widetilde{T}_{\pi^0,2}^b(L_n)}{\log
L_n} \leq\frac{1}{K_{i}(\underline{\underline{\theta}})},
\text{ for all } i\in D(\underline{\underline{\theta}}), i\in b, b\notin s(\underline{\underline{\theta}})\text{ and}\nonumber\\
&&\limsup_{n\rightarrow
\infty}\frac{E_{\underline{\underline{\theta}}}\widetilde{T}_{\pi^0,2}^b(L_n)}{\log
L_n}=0, \text{ for all } i\notin
D(\underline{\underline{\theta}}),i\in b, b\in
s(\underline{\underline{\theta}}) .\nonumber
\end{eqnarray}

{\bf Proof} We decompose $\widetilde{T}_{\pi^0,2}^b(L_n)$
as follows:

\begin{multline}
\widetilde{T}_{\pi^0,2}^b(L_n)=\sum_{t=2}^{L_n}1\{ \pi_{t}^{0}=b,
b(\hat{\underline{\underline{\theta}}}^{t})\in
s(\underline{\underline{\theta}}),
u_{i}(\hat{\underline{\underline{\theta}}}^{t},\underline{\theta}_{i}^{'})=u_{\alpha^*}(\hat{\underline{\underline{\theta}}}^{t}) \}\nonumber\\
=\sum_{t=2}^{L_n}1\{ \pi_{t}^{0}=b,
b(\hat{\underline{\underline{\theta}}}^{t})\in
s(\underline{\underline{\theta}}),
u_{i}(\hat{\underline{\underline{\theta}}}^{t},\underline{\theta}_{i}^{'})=u_{\alpha^*}(\hat{\underline{\underline{\theta}}}^{t}),
u_{i}(\hat{\underline{\underline{\theta}}}^{t},\underline{\theta}_{i}^{'})> z^* (\underline{\underline{\theta}})-\epsilon \} \nonumber\\
+\sum_{t=2}^{L_n}1\{ \pi_{t}^{0}=b,
b(\hat{\underline{\underline{\theta}}}^{t})\in
s(\underline{\underline{\theta}}),
u_{i}(\hat{\underline{\underline{\theta}}}^{t},\underline{\theta}_{i}^{'})=u_{\alpha^*}(\hat{\underline{\underline{\theta}}}^{t}),
u_{i}(\hat{\underline{\underline{\theta}}}^{t},\underline{\theta}_{i}^{'})\leq
z^* (\underline{\underline{\theta}})-\epsilon \}. \nonumber
\end{multline}

From the relation between the two indices $u_i$ and $J_i$ we have
that
\begin{gather}
\sum_{t=2}^{L_n}1\{ \pi_{t}^{0}=b,
b(\hat{\underline{\underline{\theta}}}^{t})\in
s(\underline{\underline{\theta}}),
u_{i}(\hat{\underline{\underline{\theta}}}^{t},\underline{\theta}_{i}^{'})=u_{\alpha^*}(\hat{\underline{\underline{\theta}}}^{t}),
 u_{i}(\hat{\underline{\underline{\theta}}}^{t},\underline{\theta}_{i}^{'})> z^* (\underline{\underline{\theta}})-\epsilon \} \nonumber\\
\leq\sum_{t=2}^{L_n}1\{ \pi_{t}^{0}=b,
b(\hat{\underline{\underline{\theta}}}^{t})\in
s(\underline{\underline{\theta}}),
u_{i}(\hat{\underline{\underline{\theta}}}^{t},\underline{\theta}_{i}^{'})=u_{\alpha^*}(\hat{\underline{\underline{\theta}}}^{t}),
 J_{i}(\hat{\underline{\underline{\theta}}}^{t},\epsilon)<\frac{\log
S_{\pi^0}(t-1)}{T_{\pi^0}^{i}(S_{\pi^0}(t-1))} \} \nonumber\\
=\sum_{t=2}^{L_n}1\{ \pi_{t}^{0}=b,
b(\hat{\underline{\underline{\theta}}}^{t})\in
s(\underline{\underline{\theta}}),
u_{i}(\hat{\underline{\underline{\theta}}}^{t},\underline{\theta}_{i}^{'})=u_{\alpha^*}(\hat{\underline{\underline{\theta}}}^{t}) ,\nonumber\\
J_{i}(\hat{\underline{\underline{\theta}}}^{t},\epsilon)<\frac{\log
S_{\pi^0}(t-1)}{T_{\pi^0}^{i}(S_{\pi^0}(t-1))},
 J_{i}(\hat{\underline{\underline{\theta}}}^{t},\epsilon)>J_{i}(\underline{\underline{\theta}},\epsilon)-\delta \} \nonumber\\
+\sum_{t=2}^{L_n}1\{ \pi_{t}^{0}=b,
b(\hat{\underline{\underline{\theta}}}^{t})\in
s(\underline{\underline{\theta}}),
u_{i}(\hat{\underline{\underline{\theta}}}^{t},\underline{\theta}_{i}^{'})=u_{\alpha^*}(\hat{\underline{\underline{\theta}}}^{t}), \nonumber\\
J_{i}(\hat{\underline{\underline{\theta}}}^{t},\epsilon)<\frac{\log
S_{\pi^0}(t-1)}{T_{\pi^0}^{i}(S_{\pi^0}(t-1))},
 J_{i}(\hat{\underline{\underline{\theta}}}^{t},\epsilon)\leq J_{i}(\underline{\underline{\theta}},\epsilon)-\delta \} \nonumber\\
\leq\sum_{t=2}^{L_n}1\{ \pi_{t}^{0}=b,
b(\hat{\underline{\underline{\theta}}}^{t})\in
s(\underline{\underline{\theta}}),
u_{i}(\hat{\underline{\underline{\theta}}}^{t},\underline{\theta}_{i}^{'})=u_{\alpha^*}(\hat{\underline{\underline{\theta}}}^{t}),
 T_{\pi^0}^{i}(S_{\pi^0}(t-1))<\frac{\log L_n}{J_{i}(\underline{\underline{\theta}},\epsilon)-\delta}\} \nonumber\\
+\sum_{t=2}^{L_n}1\{ \pi_{t}^{0}=b,
b(\hat{\underline{\underline{\theta}}}^{t})\in
s(\underline{\underline{\theta}}),
u_{i}(\hat{\underline{\underline{\theta}}}^{t},\underline{\theta}_{i}^{'})=u_{\alpha^*}(\hat{\underline{\underline{\theta}}}^{t}),
J_{i}(\hat{\underline{\underline{\theta}}}^{t},\epsilon)\leq
J_{i}(\underline{\underline{\theta}},\epsilon)-\delta \}. \nonumber
\end{gather}

Now, the first sum of the last inequality for $c=\frac{\log
L_n}{J_{i}(\underline{\underline{\theta}},\epsilon)-\delta}$ and $s$
integer is equal to
\begin{eqnarray}
&&\sum_{t=2}^{L_n}1\{ \pi_{t}^{0}=b,
b(\hat{\underline{\underline{\theta}}}^{t})\in
 s(\underline{\underline{\theta}}),u_{i}(\hat{\underline{\underline{\theta}}}^{t},\underline{\theta}_{i}^{'})=
 u_{\alpha^*}(\hat{\underline{\underline{\theta}}}^{t}), T_{\pi^0}^{i}(S_{\pi^0}(t-1))<c\}\nonumber\\
&\leq& \sum_{t=2}^{L_n}1\{ \pi_{t}^{0}=b,T_{\pi^0}^{i}(S_{\pi^0}(t-1))<c\}\nonumber\\
&=&\sum_{t=2}^{L_n}\sum_{s=0}^{\lfloor c/m_{i}^{b}\rfloor}1\{
\pi_{t}^{0}=
b,T_{\pi^0}^{i}(S_{\pi^0}(t-1))= s\, m_{i}^{b}+m_{i}\}\nonumber\\
&=&\sum_{s=0}^{\lfloor c/m_{i}^{b}\rfloor}\sum_{t=2}^{L_n}1\{
\pi_{t}^{0}=
b,T_{\pi^0}^{i}(S_{\pi^0}(t-1))= s\, m_{i}^{b}+m_{i}\}\nonumber\\
&\leq& \lfloor c/m_{i}^{b}\rfloor +1\nonumber\\
&\leq& \frac{c}{m_{i}^{b}} +1=\frac{\log L_n}{m_{i}^{b}(J_{i}
(\underline{\underline{\theta}},\epsilon)-\delta)}+1.\nonumber
\end{eqnarray}

Thus,
\begin{eqnarray}
&&E_{\underline{\underline{\theta}}}\sum_{t=2}^{L_n}1\{
\pi_{t}^{0}=b, b(\hat{\underline{\underline{\theta}}}^{t})\in
s(\underline{\underline{\theta}}),
u_{i}(\hat{\underline{\underline{\theta}}}^{t},\underline{\theta}_{i}^{'})=u_{\alpha^*}(\hat{\underline{\underline{\theta}}}^{t}),
 T_{\pi^0}^{i}(S_{\pi^0}(t-1))<\frac{\log L_n}{J_{i}(\underline{\underline{\theta}},\epsilon)-\delta}\}\nonumber\\
&&\leq \frac{\log
L_n}{m_{i}^{b}(J_{i}(\underline{\underline{\theta}},\epsilon)-\delta)}+1.
\label{as8}
\end{eqnarray}

Furthermore,
\begin{eqnarray}
&&\sum_{t=2}^{L_n}1\{ \pi_{t}^{0}=b,
b(\hat{\underline{\underline{\theta}}}^{t})\in
s(\underline{\underline{\theta}}),u_{i}(\hat{\underline{\underline{\theta}}}^{t},\underline{\theta}_{i}^{'})=
u_{\alpha^*}(\hat{\underline{\underline{\theta}}}^{t}),
J_{i}(\hat{\underline{\underline{\theta}}}^{t},\epsilon)\leq J_{i}(\underline{\underline{\theta}},\epsilon)-\delta \}\nonumber\\
&&\leq \sum_{t=2}^{L_n}1\{
b(\hat{\underline{\underline{\theta}}}^{t})\in
s(\underline{\underline{\theta}}),
 J_{i}(\hat{\underline{\underline{\theta}}}^{t},\epsilon)\leq J_{i}(\underline{\underline{\theta}},\epsilon)-\delta \}\nonumber
\end{eqnarray}

Then from (C2) and Remark 3 we have that
\begin{eqnarray}
&&E_{\underline{\underline{\theta}}}\sum_{t=2}^{L_n}1\{
\pi_{t}^{0}=b, b(\hat{\underline{\underline{\theta}}}^{t})\in
s(\underline{\underline{\theta}}),u_{i}(\hat{\underline{\underline{\theta}}}^{t},
\underline{\theta}_{i}^{'})=u_{\alpha^*}(\hat{\underline{\underline{\theta}}}^{t}),
J_{i}(\hat{\underline{\underline{\theta}}}^{t},
\epsilon)\leq J_{i}(\underline{\underline{\theta}},\epsilon)-\delta \}\nonumber\\
&&\leq o(\log L_n). \label{as9}
\end{eqnarray}

Now we have that
$u_{i}(\hat{\underline{\underline{\theta}}}^{t},\underline{\theta}_{i}^{'})=
u_{\alpha^*}(\hat{\underline{\underline{\theta}}}^{t})>u_{s}(\hat{\underline{\underline{\theta}}}^{t},\underline{\theta}_{s}^{'})$
for any population $s$ which is contained in an optimal BFS of
$\underline{\underline{\theta}}$. Now let
$b(\hat{\underline{\underline{\theta}}}^{t})=(r,s)$ and obviously
$b=(i,s)$, thus we can show the following inequalities
\begin{eqnarray}
&&\sum_{t=2}^{L_n}1\{ \pi_{t}^{0}=b,
b(\hat{\underline{\underline{\theta}}}^{t})\in
s(\underline{\underline{\theta}}),u_{i}(\hat{\underline{\underline{\theta}}}^{t},\underline{\theta}_{i}^{'})=
u_{\alpha^*}(\hat{\underline{\underline{\theta}}}^{t}),
u_{i}(\hat{\underline{\underline{\theta}}}^{t},
\underline{\theta}_{i}^{'})\leq z^*
(\underline{\underline{\theta}})-\epsilon \}
\nonumber\\
&&\leq\sum_{t=2}^{L_n}1\{
u_{s}(\hat{\underline{\underline{\theta}}}^{t},\underline{\theta}_{s}^{'})\leq
z^* (\underline{\underline{\theta}})-\epsilon \}
\nonumber\\
&&\leq\sum_{t=2}^{L_n}1\{
u_{s}(\hat{\underline{\underline{\theta}}}^{j},\underline{\theta}_{s}^{'})\leq
 z^* (\underline{\underline{\theta}})-\epsilon, \text{ for some }j\leq S_{\pi^0}(t-1) \}\nonumber\\
&&=\sum_{t=2}^{L_n}1\{
|\hat{\underline{\theta}}_{s}^{j}-\underline{\theta}_s|>\xi,\text{
for some }j\leq S_{\pi^0}(t-1) \}.\nonumber
\end{eqnarray}

Thus
\begin{eqnarray}
&&E_{\underline{\underline{\theta}}}\sum_{t=2}^{L_n}1\{
\pi_{t}^{0}=b, b(\hat{\underline{\underline{\theta}}}^{t})\in
s(\underline{\underline{\theta}}),
u_{i}(\hat{\underline{\underline{\theta}}}^{t},\underline{\theta}_{i}^{'})=
u_{\alpha^*}(\hat{\underline{\underline{\theta}}}^{t}),
u_{i}(\hat{\underline{\underline{\theta}}}^{t},\underline{\theta}_{i}^{'})\leq z^* (\underline{\underline{\theta}})-\epsilon \}\nonumber\\
&&\leq o(\log L_n), \label{as10}
\end{eqnarray}
because
\begin{eqnarray}
&&P_{\underline{\theta}_s}^{\pi^0}(|\hat{\underline{\theta}}_{s}^{j}-\underline{\theta}_s|>\xi,\text{ for some }j\leq t)\nonumber\\
&&\leq
\sum_{j=1}^{t}P_{\underline{\theta}_s}^{\pi^0}(|\hat{\underline{\theta}}_{s}^{j}-\underline{\theta}_s|>\xi)=o(1/t),\nonumber
\end{eqnarray}
since policy $\pi^0$ at any block $t$ chooses
$b(\hat{\underline{\underline{\theta}}}^{t})=(r,s)$ when
$\widetilde{T}_{\pi^0}^{b(\hat{\underline{\underline{\theta}}}^{t})}(t)\geq
\tau(t-1)$.

Finally, it follows from \eqref{as8}, \eqref{as9} and \eqref{as10}
that
\begin{equation}
E_{\underline{\underline{\theta}}}\widetilde{T}_{\pi^0}^{b}(L_n)\leq
 \frac{\log L_n}{m_{i}^{b}(J_{i}(\underline{\underline{\theta}},\epsilon)-\delta)} +1 +o(\log L_n)+o(\log L_n).\nonumber
\end{equation}

Now from the definition of
$J_{i}(\underline{\underline{\theta}},\epsilon)$ and (C1) we have
that
\begin{equation}
\lim_{\epsilon\rightarrow
0}J_{i}(\underline{\underline{\theta}},\epsilon)=
K_i(\underline{\underline{\theta}}),\text{ for }i\in
D(\underline{\underline{\theta}})\text{  and  }
\lim_{\epsilon\rightarrow
0}J_{i}(\underline{\underline{\theta}},\epsilon)=\infty,\text{ for
}i\notin D(\underline{\underline{\theta}}).\nonumber
\end{equation}

Thus
\begin{eqnarray}
&&\limsup_{n\rightarrow
\infty}\frac{E_{\underline{\underline{\theta}}}\widetilde{T}_{\pi^0,2}^{b}(L_n)}{\log
L_n} \leq\frac{1}{K_{i}(\underline{\underline{\theta}})}, \text{ for
all } i\in D(\underline{\underline{\theta}}), i\in b, b\notin
s(\underline{\underline{\theta}})
\text{ and}\nonumber\\
&&\limsup_{n\rightarrow
\infty}\frac{E_{\underline{\underline{\theta}}}\widetilde{T}_{\pi^0,2}^{b}(L_n)}{\log
L_n}=0, \text{ for all } i\notin
D(\underline{\underline{\theta}}),i\in b, b\in
s(\underline{\underline{\theta}}) .\nonumber
\end{eqnarray}
$\square$

\vspace{1cm}For the next Lemma, let $0<\varepsilon<\{ z^*
(\underline{\underline{\theta}})- \max_{b\notin
s(\underline{\underline{\theta}})}z^{b(\underline{\underline{\theta}})}
\}/2$ and $c$ a positive integer. Then we define for $r=0,1,2,...$

\begin{equation}
A_r=\bigcap_{1\leq j\leq |F|} \{ \max_{\tau c^{r-1}\leq l-1\leq
c^{r+1}}|z^{\widetilde{b}_{j}(\hat{\underline{\underline{\theta}}}^{l})}-
z^{\widetilde{b}_{j}(\underline{\underline{\theta}})}|\leq\varepsilon
\} \text{ and } \nonumber
\end{equation}
\begin{equation}
B_r=\bigcap_{b_{\alpha} \in s(\underline{\underline{\theta}})}\{
z^{b_{\alpha}(\hat{\underline{\underline{\theta}}}^{i},\underline{\theta}_{\alpha}^{'})}\geq
z^* (\underline{\underline{\theta}})-\varepsilon, \text{ for all
}1\leq i\leq \tau(l-1)\text{ and }c^{r-1}\leq l-1 \leq c^{r+1}
\},\nonumber
\end{equation}
where $0<\tau<1/|F|$ is the same as in the $\pi^0$.

\vspace{1cm}{\bf Lemma 5}
Under conditions (C2),(C3)\\
 \emph{\textbf{(i)}}
$P_{\underline{\underline{\theta}}}^{\pi^0}(\overline{A}_r)=o(c^{-r})$,
$P_{\underline{\underline{\theta}}}^{\pi^0}(\overline{B}_r)=o(c^{-r})$.\\
Moreover, if $c>1/(1-|F|\tau)$ and $r\geq r_0$ then\\
\emph{\textbf{(ii)}} on $A_r\cap B_r$,
$b(\hat{\underline{\underline{\theta}}}^{l})\in
s(\underline{\underline{\theta}})$
for all $c^{r-1}\leq l-1\leq c^{r+1}$.\\
\emph{\textbf{(iii)}}
$E_{\underline{\underline{\theta}}}\widetilde{T}_{\pi^0,1}^{b}(L_n)=
\sum_{t=2}^{L_n}P_{\underline{\underline{\theta}}}^{\pi^0}
(b(\hat{\underline{\underline{\theta}}}^{t})\notin
s(\underline{\underline{\theta}}))=o(\log L_n)$.

{\bf Proof} (i) We have that from (C2)
\begin{equation}
P_{\underline{\underline{\theta}}}^{\pi^0}(\max_{\tau c^{r-1}\leq
l-1\leq
c^{r+1}}|z^{\widetilde{b}_{j}(\hat{\underline{\underline{\theta}}}^{l})}-
z^{\widetilde{b}_{j}(\underline{\underline{\theta}})}|>\varepsilon)=
o(c^{-r}), \ 1\leq j\leq |F|\nonumber
\end{equation}
holds for the sample mean of the estimates
$\hat{\underline{\underline{\theta}}}^{l}=
\frac{\hat{\underline{\underline{\theta}}}^{1} +...+
\hat{\underline{\underline{\theta}}}^{l}}{l-1}$ thus it follows that
$P_{\underline{\underline{\theta}}}^{\pi^0}(\overline{A}_r)=o(c^{-r})$.

Now let $q$ be the smallest positive integer such that $\lfloor
c^{r-1}/\tau^q\rfloor \geq c^{r+1}$. For $t=0,...,q$ and
$l_t=\lfloor c^{r-1}/\tau^t\rfloor$ we define the sets
\begin{equation}
Q_t=\bigcap_{b_{\alpha}\in s(\underline{\underline{\theta}})}
\left\{
z^{b_{\alpha}(\hat{\underline{\underline{\theta}}}^{i},\underline{\theta}_{\alpha}^{'})}\geq
z^* (\underline{\underline{\theta}})-\varepsilon, \text{ for all
}1\leq i\leq l_t \right\}.\nonumber
\end{equation}

Then by (C3),
\begin{equation}
P_{\underline{\underline{\theta}}}^{\pi^0}(\overline{Q}_t)= o(1/l_t)
= o(c^{-r}) \text{ for } t=0,...,q. \label{lem5.4}
\end{equation}

Now given that $c^{r-1}\leq l-1\leq c^{r+1}$ and $1\leq i \leq \tau
(l-1)$, there exists $t\in \{ 0,...,q \}$ such that $l_{t+1}>l-1\geq
l_t \geq i$ and therefore for every fix $b_{\alpha}$ we have that
\begin{equation}
z^{b_{\alpha}(\hat{\underline{\underline{\theta}}}^{l},\underline{\theta}_{\alpha}^{'})}
\geq
z^{b_{\alpha}(\hat{\underline{\underline{\theta}}}^{l_{t}},\underline{\theta}_{\alpha}^{'})}\geq
z^* (\underline{\underline{\theta}})-\varepsilon. \nonumber
\end{equation}
for every $b_{\alpha}\in s(\underline{\underline{\theta}})$ on the
event $\bigcap_{0\leq t \leq q}Q_t$. Thus, because of $B_r \supset
\bigcap_{0\leq t \leq q}Q_t$ and \eqref{lem5.4} we have that
$P_{\underline{\underline{\theta}}}^{\pi^0}(\overline{B}_r)=o(c^{-r})$.

(ii) Let
$V_{s(\underline{\underline{\theta}})}^{\pi^0}(l)=\sum_{b\in
s(\underline{\underline{\theta}})}\widetilde{T}_{\pi^0}^{b}(l)$ be
the number of times that $\pi^0$ samples from
$s(\underline{\underline{\theta}})$ up to $l$ sampling block.

We note that
\begin{equation}
\max_{b\in
s(\underline{\underline{\theta}})}\widetilde{T}_{\pi^0}^{b}(l) \geq
\frac{V_{s(\underline{\underline{\theta}})}^{\pi^0}(l)}{
|s(\underline{\underline{\theta}})|}, \label{lem5.1}
\end{equation}
where $|s|$ denotes the number of elements of $s$.

Consider that at any block $l$ and $c^{r-1}\leq l-1\leq c^{r+1}$, we
have that $u_{\alpha^*}(\hat{\underline{\underline{\theta}}}^{l})
\in s(\underline{\underline{\theta}})$, and
$u_{\alpha^*}(\hat{\underline{\underline{\theta}}}^{l})$ corresponds
to an optimal BFS
$b_{\alpha^*}(\hat{\underline{\underline{\theta}}}^{l})$. Then if
$b(\hat{\underline{\underline{\theta}}}^{l})\in
s(\underline{\underline{\theta}})$ we have the requested. Now, let
assume that $b(\hat{\underline{\underline{\theta}}}^{l})\notin
s(\underline{\underline{\theta}})$, and we have that
$b_{\alpha^*}(\hat{\underline{\underline{\theta}}}^{l}) \in
s(\underline{\underline{\theta}})$ which means that on $A_r\cap B_r$
the policy $\pi^0$ chooses from $s(\underline{\underline{\theta}})$.

Then since
$\widetilde{T}_{\pi^0}^{b(\hat{\underline{\underline{\theta}}}^{l})}
(l)\geq \tau (l-1)$,
\begin{equation}
z^{b(\hat{\underline{\underline{\theta}}}^{l})} \leq \max_{b\notin
s(\underline{\underline{\theta}})}
z^{b(\underline{\underline{\theta}})} + \varepsilon <
z^{*}(\underline{\underline{\theta}}) - \varepsilon \text{ on } A_r.
\nonumber
\end{equation}
In the case where
$\widetilde{T}_{\pi^0}^{b_{\alpha^*}(\hat{\underline{\underline{\theta}}}^{l})}
(l)\geq \tau (l-1)$, we have on the event $A_r$
\begin{equation}
z^{*}(\underline{\underline{\theta}}) - \varepsilon \leq
z^{b_{\alpha^*}(\hat{\underline{\underline{\theta}}}^{l})}.\nonumber
\end{equation}
In the other case where
$\widetilde{T}_{\pi^0}^{b_{\alpha^*}(\hat{\underline{\underline{\theta}}}^{l})}
(l)< \tau (l-1)$, we have on the event $B_r$
\begin{equation}
z^{*}(\underline{\underline{\theta}}) - \varepsilon \leq
z^{b_{\alpha^*}(\hat{\underline{\underline{\theta}}}^{l})}.\nonumber
\end{equation}

On the event $A_r\cap B_r$, since $\pi^0$ employs from
$s(\underline{\underline{\theta}})$ at block $l$ and $c^{r-1}\leq
l-1\leq c^{r+1}$, and since $c>1/(1-|F|\tau)$ it follows that
\begin{equation}
V_{s(\underline{\underline{\theta}})}^{\pi^0}(l) \geq \frac{|
s(\underline{\underline{\theta}})|}{|F|} (l-1-c^{r-1}-2|F|)
> (
|s(\underline{\underline{\theta}})|) \tau (l-1) \label{lem5.2}
\end{equation}
for all $c^{r-1}\leq l-1\leq c^{r+1}$ and $r\geq r_0$.

From \eqref{lem5.1} and \eqref{lem5.2}, we obtain on $A_r\cap B_r$
\begin{equation}
\max_{b\in
s(\underline{\underline{\theta}})}\widetilde{T}_{\pi^0}^{b}(l) >
\tau (l-1) \label{lem5.3}
\end{equation}
for all $c^{r-1}\leq l-1\leq c^{r+1}$ if $r\geq r_0$.

We note that for $r\geq r_0$ and $c^{r-1}\leq l-1\leq c^{r+1}$, on
the event $A_r\cap B_r$,
\begin{eqnarray}
&&\max\{ z^b : \widetilde{T}_{\pi^0}^{b}(l) \geq \tau (l-1) \text{
and }
b \notin s(\underline{\underline{\theta}}) \}\nonumber\\
&& \leq \max_{b \notin s(\underline{\underline{\theta}})}z^b +
\varepsilon
< z^*(\underline{\underline{\theta}}) - \varepsilon\nonumber\\
&&\leq\min\{ z^b : \widetilde{T}_{\pi^0}^{b}(l) \geq \tau (l-1)
\text{ and } b \in s(\underline{\underline{\theta}}) \}\nonumber
\end{eqnarray}
the last set is nonempty because of \eqref{lem5.3}. Hence
$b(\hat{\underline{\underline{\theta}}}^{l})\in
s(\underline{\underline{\theta}})$ for all $c^{r-1}\leq l-1\leq
c^{r+1}$ on the event $A_r\cap B_r$ if $r\geq r_0$.

(iii) Let $c>1/(1-|F|\tau)$. Then it follows from (i) and (ii) that
for $r\geq r_0$ and $c^{r-1}\leq t-1\leq c^{r+1}$,
\begin{equation}
P_{\underline{\underline{\theta}}}^{\pi^0}
(b(\hat{\underline{\underline{\theta}}}^{t})\notin
s(\underline{\underline{\theta}})) \leq
P_{\underline{\underline{\theta}}}^{\pi^0}(\overline{A}_r) +
P_{\underline{\underline{\theta}}}^{\pi^0}(\overline{B}_r) =
o(c^{-r})\nonumber
\end{equation}
and therefore
\begin{equation}
\sum_{c^{r-1}\leq t-1\leq c^{r+1}}
P_{\underline{\underline{\theta}}}^{\pi^0}
(b(\hat{\underline{\underline{\theta}}}^{t})\notin
s(\underline{\underline{\theta}})) =o(1).\nonumber
\end{equation}
Hence,
\begin{equation}
\sum_{t=2}^{L_n} P_{\underline{\underline{\theta}}}^{\pi^0}
(b(\hat{\underline{\underline{\theta}}}^{t})\notin
s(\underline{\underline{\theta}})) =o(\log L_n).\nonumber
\end{equation}
$\square$

\end{document}